\documentclass[preprint,12pt,authoryear]{elsarticle}
\usepackage{cite}
\usepackage{amsmath,amssymb,amsfonts}
\usepackage{algorithmic}
\usepackage{graphicx}
\usepackage{textcomp}

\usepackage{xcolor}
\usepackage{comment}
\usepackage{amsmath}
\usepackage{subfigure}
\usepackage{algorithm2e}

\usepackage{siunitx}
\usepackage[utf8]{inputenc} 
\usepackage[T1]{fontenc}    
\usepackage{hyperref}       
\usepackage{url}            
\usepackage{booktabs}       
\usepackage{amsfonts}       
\usepackage{nicefrac}       
\usepackage{microtype}      
\usepackage{float}          
\usepackage{amsmath}
\usepackage{float}
\usepackage{booktabs}
\usepackage{graphicx}
\usepackage{subcaption}
\usepackage{enumitem}
\usepackage{xcolor}
\usepackage{hyperref}
\usepackage{multirow}
\usepackage{tabularx}
\usepackage{multirow}
\usepackage{makecell}
\usepackage{booktabs}
\usepackage{colortbl}
\usepackage{rotating}
\usepackage{array}
\usepackage{adjustbox}
\usepackage{wrapfig}
\usepackage{url}
\usepackage{cite}

\SetKwInput{KwInput}{Input}
\SetKwInput{KwOutput}{Output}

\usepackage{color}
\usepackage{tabularray}
\usepackage{listings}
\usepackage{xcolor}
\usepackage{threeparttable}
\usepackage{listings}
\usepackage{xcolor}
\usepackage{minted}
\usepackage{xcolor}

\usepackage{tcolorbox} 

\lstdefinestyle{python}{
    language=Python,
    backgroundcolor=\color{lightgray!20}, 
    basicstyle=\ttfamily\footnotesize, 
    keywordstyle=\color{blue}, 
    stringstyle=\color{red}, 
    commentstyle=\color{green!50!black}, 
    identifierstyle=\color{purple}, 
    morekeywords={def}, 
    morestring=[b]", 
    morestring=[b]', 
    breaklines = true
}

\usepackage{tcolorbox}
\newtcolorbox[auto counter]{promptbox}[2][]{
    colback=gray!5!white,
    colframe=gray!75!black,
    fonttitle=\bfseries,
    title=Prompt~\thetcbcounter: #2,  
    arc=3pt,
    boxrule=0.5pt,
    left=6pt, right=6pt, top=6pt, bottom=6pt,
    #1 
}

\journal{International Journal of Forecasting}

\begin{document}

\begin{frontmatter}

\title{Bridging AI and Energy Forecasting: An Autonomous Workflow with Customized Toolkit}

\author[inst1,inst2]{Zhixian Wang}
\author[inst3]{Leandro Von Krannichfeldt}
\author[inst4]{Qingsong Wen}
\author[inst5]{Chaoli Zhang}
\author[inst6]{Liang Sun}
\author[inst7]{Shirui Pan}
\author[inst1,inst2]{Yi Wang\footnote{Corresponding author: yiwang@eee.hku.hk}}

\affiliation[inst1]{organization={Department of Electrical and Electronic Engineering, The University of Hong Kong},
            postcode={999077}, 
            state={Hong Kong SAR},
            country={China}}

\affiliation[inst2]{organization={Shenzhen Institute of Research and Innovation,
The University of Hong Kong (Shenzhen)},
            postcode={518057}, 
            state={Shenzhen},
            country={China}}

\affiliation[inst3]{organization={EPF Lausanne},
            city={Lausanne},
            country={Switzerland}}

\affiliation[inst4]{organization={Squirrel Ai Learning},
           city={Bellevue, WA},
           country={USA}}

\affiliation[inst5]{organization={Zhejiang Normal University},
           city={Jinhua},
           country={China}}
\affiliation[inst6]{organization={DAMO Academy, Alibaba Group (U.S.) Inc.},
            city={Bellevue},
            postcode={WA 98004}, 
            country={USA}}
\affiliation[inst7]{organization={Griffith University},
           city={Brisbane},
           country={Australia}}

\begin{abstract}
Energy forecasting is crucial for the power grid, but fundamentally different from general time series analysis: it highly relies on covariates like meteorological factors, and its goals must align with actual power grid operations, such as risk assessment and system reliability. In order to bridge the huge gap between advanced machine learning forecasting models and actual power grid demand, this paper proposes an autonomous forecasting workflow based on LLMs. As a virtual analyst, the agent replaces tedious manual adjustments by autonomously analyzing data features, dynamically orchestrating optimal forecasting pipelines, and generating actionable analysis reports for decision-makers. The carefully arranged pipeline directly addresses the demand of the power grid through probabilistic forecasting for uncertainty quantification. Furthermore, the framework acts as an automated testbed, enabling seamless A/B testing of specific algorithmic plugins across various architectures to evaluate their empirical effectiveness. As the foundation of the algorithm, the underlying toolkit integrates \textbf{31} advanced temporal architectures and \textbf{6} customized exogenous modules, resulting in \textbf{146} highly configurable variants. In addition, to provide experience references for the agents, we established a large-scale benchmark on \textbf{21} energy datasets and released new high-quality renewable energy datasets with meteorological factors. 

\end{abstract}




\end{frontmatter}

\section{Introduction}
Time series data are becoming ubiquitous in numerous real-world applications \cite{wen2022robust,lai2021revisiting,zhou2022film,wang2018review}. Among them, energy forecasting is crucial for maintaining the supply and demand balance in the power system. Thanks to the development of machine learning in recent years, various methods have been developed for energy forecasting \cite{yildiz2017review, zhang2021review,aslam2021survey,benti2023forecasting, Yang2023GAM, Zhu2023eForecaster}. Although many advanced forecasting methods have emerged in the past decades, they still face significant bottlenecks in top-tier forecasting competitions and actual implementations within the energy sector \cite{hong2014global, hong2016probabilistic, hong2019global, nweye2022citylearn, farrokhabadi2022day}. On the one hand, they lack domain knowledge that is extremely important for energy forecasting, such as the impact of temperature on power load \cite{sobhani2020temperature}. On the other hand, they mainly focus on general statistical metrics and rarely consider the application characteristics of the energy sequence itself, such as the high uncertainty of renewable energy and the ultimate goals of energy forecasting, which must directly support actual grid operations like risk assessment and system reliability.

Firstly, compared with general time series, energy data are greatly affected by auxiliary variables such as temperature, making it challenging to model the dynamics accurately. Therefore, exploring the impact of auxiliary variables on energy forecasting has always been an important research direction in this field \cite{aprillia2020statistical}. For load data, temperature is considered to have a significant impact and many researchers have focused on using temperature variables to assist in constructing load forecasting models \cite{haben2019short, sobhani2020temperature,liu2023sadi}. As for renewable energy data, the impact of auxiliary variables is even more significant, such as the effect of wind speed on wind power. However, unlike load data, a common problem in the community is that we often lack meteorological variables that match the renewable energy data. As stated in \cite{menezes2020wind}, an excellent renewable energy dataset not only requires generation data but also corresponding meteorological data, which is often difficult. For this situation, we have collected and publicly released a renewable energy output dataset from a first-level administrative region with a highly developed economy, covering different types of renewable energy data such as PV, onshore wind, and offshore wind, as well as corresponding key meteorological factors such as real-time wind speed at wind turbine hubs. As for electricity price data, the auxiliary variables that can affect it are no longer just meteorological factors. Further factors, such as renewable energy output and load demand, and even economic situation, will have a very large impact \cite{heistrene2024improved}.

Secondly, deploying state-of-the-art machine learning models into actual power systems is not a simple "plug and play" process; it needs to bridge the fundamental gap between mathematical optimization and grid demand. Traditionally, seamlessly integrating complex forecasting architectures with practical grid requirements—such as uncertainty quantification for renewable energy or evaluating algorithmic compatibility with specific operational constraints—requires exhaustive trial-and-error and heuristic manual adjustments. And this process often requires a lot of manpower and material resources. Fortunately, recent profound breakthroughs in Large Language Models (LLMs) have provided a transformative pathway to overcome this human bottleneck. Modern LLM has surpassed simple language processing and demonstrated outstanding abilities in logical reasoning, power domain knowledge synthesis\cite{zhou2025elecbench, lozano2025democratizing}, and automatic code generation. Based on this, LLM will effectively assist in the implementation of advanced forecasting methods in the actual power industry.

To truly apply the development advancements of machine learning to actual energy systems and overcome the aforementioned data characteristic and domain knowledge bottleneck, we construct an overarching automated pipeline. Firstly, we develop a comprehensive, power-centric energy forecasting toolkit. Distinct from existing general time series packages \cite{alexandrov2020gluonts, godahewa2021monash,olivares2022library_neuralforecast}, our toolkit logically splits the forecasting process into five highly customizable stages: data preprocessing, feature engineering, forecasting methods, postprocessing, and evaluation metrics. Each phase deeply incorporates power domain expertise. For instance, we embed the domain specific temperature-calendar feature engineering \cite{hong2016probabilistic} and and provide optional plug-and-play modules like dispatch cost-oriented loss functions \cite{zhang2022cost} for scenario-specific exploration. Secondly, utilizing this toolkit, we establish a large-scale energy forecasting benchmark covering 21 diverse datasets (spanning load, renewable energy output, and electricity prices). This extensive benchmark serves as a crucial empirical reference, mapping various data characteristics to the performance of different forecasting architectures. Finally, to completely automate the deployment process for future scenarios, we introduce an LLM-empowered autonomous forecasting agent. Acting as a virtual forecasting analyst, this agent leverages Retrieval-Augmented Generation (RAG) to draw upon the empirical knowledge from our established benchmark and domain literature. When presented with unseen energy data, the agent autonomously analyzes the data characteristics, logically selects and dynamically invokes the optimal forecasting pipelines from the toolkit. Moreover, the agent enables automated A/B testing across specific plugins and architectures to empirically evaluate their actual suitability. Ultimately, it executes the prediction and generates detailed, naturally readable analytical reports for grid decision-makers, comprehensively comparing expected accuracies, computational costs, and method suitabilities tailored to the new data.

\textbf{We summarize our primary contributions as follows}:

\begin{enumerate}[leftmargin=*] 
    \item \textbf{A highly customizable, energy-centric algorithm toolkit}: To serve as a solid foundation for cognitive agents, we have developed a comprehensive toolkit that integrates \textbf{31} advanced temporal architectures and \textbf{6} specialized exogenous covariate modules (spanning \textbf{146} highly configurable variants). Crucially, the toolkit narrows the gap between machine learning operations and actual grid demands by natively supporting probabilistic forecasting for risk uncertainty quantification. Additionally, it provides optional modular plugins, such as customized cost-oriented loss functions, to facilitate scenario-specific optimizations and explorations.

    \item \textbf{A large-scale benchmark archive and new renewable energy datasets}: We have established a comprehensive benchmark across \textbf{21} diverse energy datasets (load, wind, photovoltaic, and electricity prices) to serve as an essential empirical knowledge reference for the agents. Furthermore, we publicly release high-quality, newly compiled renewable energy generation datasets that are precisely aligned with real-time, localized meteorological factors. Our datasets uniquely provide renewable energy generation data at multiple spatial levels—from specific stations to larger regional aggregations—offering a highly valuable reference for researchers in the field.

    \item \textbf{An LLM-empowered autonomous forecasting framework}: We propose an innovative autonomous agent framework acting as a virtual forecasting analyst for power systems. By utilizing Retrieval-Augmented Generation (RAG) and domain knowledge, the agent autonomously analyzes new data characteristics and dynamically orchestrates optimal forecasting pipelines, drastically reducing expert manual tuning. Notably, the framework functions as an automated testbed capable of conducting autonomous comparative A/B testing to evaluate specific algorithmic modules. Ultimately, it generates detailed, decision-ready analytical reports to directly assist grid operators.
\end{enumerate}
\section{The energy forecasting tookit}
\subsection{Overview of the tookit}
\begin{figure}[htbp]
\centering
\includegraphics[width=\textwidth]{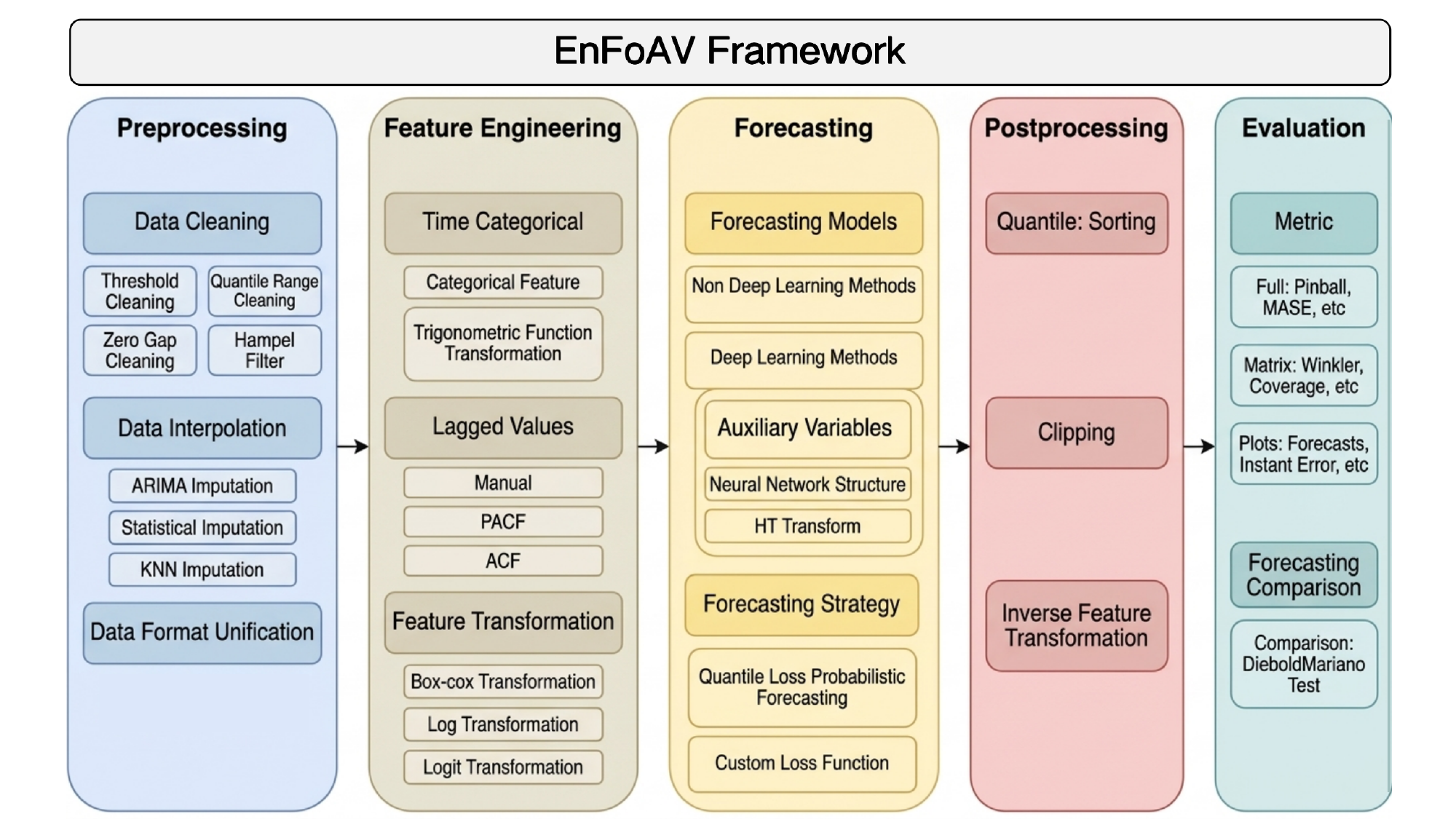}
\centering
\caption{Overview of the energy forecasting tookit.}
\label{tookit}
\end{figure}

Figure \ref{tookit} shows the overview of our toolkit. We logically divide the forecasting process into several customizable stages to address specific challenges in the power industry. First, to handle inevitable missing values collected from physical meters, the toolkit provides robust imputation methods like Kalman filtering-based ARIMA \cite{harvey1984estimating} and K-nearest neighbor algorithms \cite{garcia2010pattern}. Secondly, it offers versatile feature selection strategies, such as using past seven-day data for day-ahead tasks or applying ACF/PACF metrics to select lagged values. Regarding forecasting models, we integrate 3 non-deep learning variants and 27 deep learning architectures. Given the high uncertainty introduced by renewable energy and extreme weather \cite{haupt2019use}, we implemented corresponding probabilistic forecasting versions via quantile regression for all neural networks (detailed in Section \ref{sec3.2}). Furthermore, to respect explicit physical limitations like installed capacity bounds, dedicated post-processing methods are provided to constrain the raw forecasts. Finally, recognizing that existing holistic metrics (e.g., overall Pinball Loss) fail to provide detailed performance guidelines across different probability distributions, our toolkit introduces matrix visualization functions. This enables granular performance comparisons across various quantiles using comprehensive metrics like CalibrationError \cite{chung2021beyond}, WinklerScore \cite{barnett1973introduction}, and CoverageError (details in Appendix Section \ref{evaluation_metrics}).

\subsection{Forecasting methods, auxiliary variables modules, and cost-oriented loss function}\label{sec3.2}
In our tookit, we have implemented 3 non-deep learning methods and 28 deep learning methods including most of the state-of-art time series forecasting models. In energy forecasting, we often need to consider the future information of auxiliary variables, such as weather forecasts. Therefore, based on whether auxiliary variables can be considered and whether the future information of auxiliary variables can be considered, we have compiled these 31 forecasting methods. Table \ref{fig:Forecasting_models} summarizes the forecasting method that we implement. Among them, there are as many as 22 general time series forecasting methods, including simple forward propagation networks \cite{jain1996artificial}, LSTM networks \cite{hochreiter1997long} for sequence modeling, convolutional neural networks \cite{li2021survey} (where we use one-dimensional convolutional kernels), Transformer \cite{vaswani2017attention} networks applying attention mechanisms. In addition, we include methods that modify and combine the above neural network structures to make them more suitable for forecasting tasks, such as LSTNet \cite{lai2018modeling}, which is designed to simultaneously capture both long-term and short-term patterns of time series, WaveNet based on causal convolution \cite{oord2016wavenet}, N-BEATS that stacked into blocks using multiple linear layers \cite{oreshkin2019n}, Dlinear \cite{zeng2023transformers} which only use linear layer to model the series, Frequency improved Legendre Memory model \cite{zhou2022film}, SegRNN that based on segment-wise Iterations and parallel multi-step forecasting \cite{lin2023segrnn}, TimesNet that extend the analysis of temporal variations into the 2D space \cite{wutimesnet}, and MICN \cite{wang2023micn} that a multi-scale branch structure is adopted to model different potential patterns separately. In recent years, Transformer-based forecasting models have been widely applied. Therefore, we also include related models like Informer \cite{zhou2021informer}, Autoformer \cite{wu2021autoformer}, Fedformer \cite{zhou2022fedformer}, iTransformer \cite{liuitransformer}, NSTransformer \cite{liu2022non}, PatchTST \cite{nietime}, and Reformer \cite{kitaev2019reformer}. Although these Transformer-based models have achieved good results, their computational complexity is often very high. In this case, we also include some MLP-based forecasting models like Timemixer \cite{wangtimemixer}, Tsmixer \cite{chen2023tsmixer}, and FreTS \cite{yi2024frequency} that achieve good results with less computational burden.

However, in addition to non-deep learning methods like K-nearest neighbor algorithm \cite{hastie2009elements}, Random Forest \cite{meinshausen2006quantile}, and Extreme Random Tree \cite{geurts2006extremely} that do not consider the characteristics of time series and treat all features as equal, there are only 6 forecasting methods that can consider auxiliary variables to a certain extent \cite{wang2024timexer,oreshkin2021n,sprangers2023parameter,lim2021temporal,daslong,chen2023tsmixer}. This comparison shows that there is still considerable room for exploration in time series forecasting considering auxiliary variables, which hinders the application of deep learning models in real energy forecasting scenarios. Based on this situation, in order to fully apply the existing general time series forecasting technology to the energy forecasting scenario, we implement 4 common neural network structures and a proven effective time series feature processing module called Mixer \cite{chen2023tsmixer} in our tookit, and combining them with the existing general time series model for the task of time series forecasting considering auxiliary variables. 

\begin{table}[htbp]
\centering
\caption{Time series forecasting model archive in our tookit.}
\renewcommand{\arraystretch}{1.2}
\resizebox{0.85\textwidth}{!}{
\begin{tabular}{c|c|c|c|c}
\hline
\textbf{Category}    & \textbf{Model} & \textbf{Type$^\S$} & \textbf{Auxiliary variables} & \textbf{Future auxiliary variables} \\ \hline
\multirow{20}{*}{I}  & MLP \cite{jain1996artificial}            & D             &              $\times$                &               $\times$                      \\
                     & LSTM \cite{hochreiter1997long}          & D             &               $\times$               &               $\times$                      \\
                     & CNN \cite{li2021survey}            & D             &              $\times$                &              $\times$                       \\
                     & Transformer \cite{vaswani2017attention}    & D             &                $\times$              &             $\times$                        \\
                     & LSTNet   \cite{lai2018modeling}      & D             &               $\times$               &             $\times$                        \\
                     & WaveNet \cite{oord2016wavenet}        & D             &            $\times$                  &              $\times$                       \\
                     & N-BEATS \cite{oreshkin2019n}         & D             &                $\times$              &             $\times$                        \\
                     & Informer \cite{zhou2021informer}      & D             &                 $\times$             &             $\times$                        \\
                     & Autoformer \cite{wu2021autoformer}    & D             &                  $\times$            &             $\times$                        \\
                     & Fedformer  \cite{zhou2022fedformer}    & D             &                  $\times$            &             $\times$                        \\
                     & DLinear   \cite{zeng2023transformers}   & D             &                  $\times$            &             $\times$                        \\
                     & FiLM \cite{zhou2022film}           & D             &               $\times$               &              $\times$                       \\
                     & iTransformer \cite{liuitransformer}   & D             &            $\times$                  &            $\times$                         \\
                     & NSTransformer \cite{liu2022non}  & D             &            $\times$                  &             $\times$                        \\
                     & PatchTST \cite{nietime}      & D             &             $\times$                 &             $\times$                        \\
                     & SegRNN \cite{lin2023segrnn}         & D             &           $\times$                   &             $\times$                        \\
                     & TimeMixer \cite{wangtimemixer}      & D             &            $\times$                  &              $\times$                       \\
                     & TimesNet  \cite{wutimesnet}     & D             &          $\times$                    &             $\times$                        \\
                     & Tsmixer \cite{chen2023tsmixer}        & D             &            $\times$                  &             $\times$                        \\
                     & FreTS \cite{yi2024frequency}          & D             &            $\times$                  &             $\times$                        \\
                     & Reformer \cite{kitaev2019reformer}       & D             &            $\times$                  &             $\times$                        \\
                     & MICN \cite{wang2023micn}           & D             &            $\times$                  &              $\times$                       \\ \hline
II                   & TimeXer\textsuperscript{\dag} \cite{wang2024timexer}        & D             &             $\checkmark$                 &                       $\times$              \\ \hline
\multirow{8}{*}{III} & N-BEATSX \cite{oreshkin2021n}        & D             &             $\checkmark$                 &                  $\checkmark$                   \\
                     & BiTCN \cite{sprangers2023parameter}          & D             &              $\checkmark$                &                 $\checkmark$                    \\
                     & TFT \cite{lim2021temporal}            & D             &              $\checkmark$                &                $\checkmark$                     \\
                     & TiDE \cite{daslong}          & D             &              $\checkmark$                &                 $\checkmark$                    \\
                     & TsmixerEXT \cite{chen2023tsmixer}     & D             &             $\checkmark$                 &                $\checkmark$                     \\
                     & KNNR\textsuperscript{*} \cite{hastie2009elements}           & N             &              $\checkmark$                &                $\checkmark$                     \\
                     & RFR\textsuperscript{*} \cite{meinshausen2006quantile}            & N             &             $\checkmark$                 &                $\checkmark$                     \\
                     & ERT\textsuperscript{*} \cite{geurts2006extremely}          & N             &              $\checkmark$                &               $\checkmark$                      \\ \hline
\end{tabular}
}
\begin{tablenotes}
\footnotesize
\item[1] $\S$ : Here D represents deep learning methods while N represents non-deep learning methods.
\item[2] $\dag$ : TimeXer can consider auxiliary variables of varying lengths, but in implementation, it mainly considers historical auxiliary variables, so we classify it as Class II here.
\item[3] $*$: Non-deep learning methods cannot consider time series relationships. Here we input all relevant variables, including the future information, into the model.
\end{tablenotes}

\label{fig:Forecasting_models}
\end{table}

\begin{figure}[htbp]
\centering
\includegraphics[width=\textwidth]{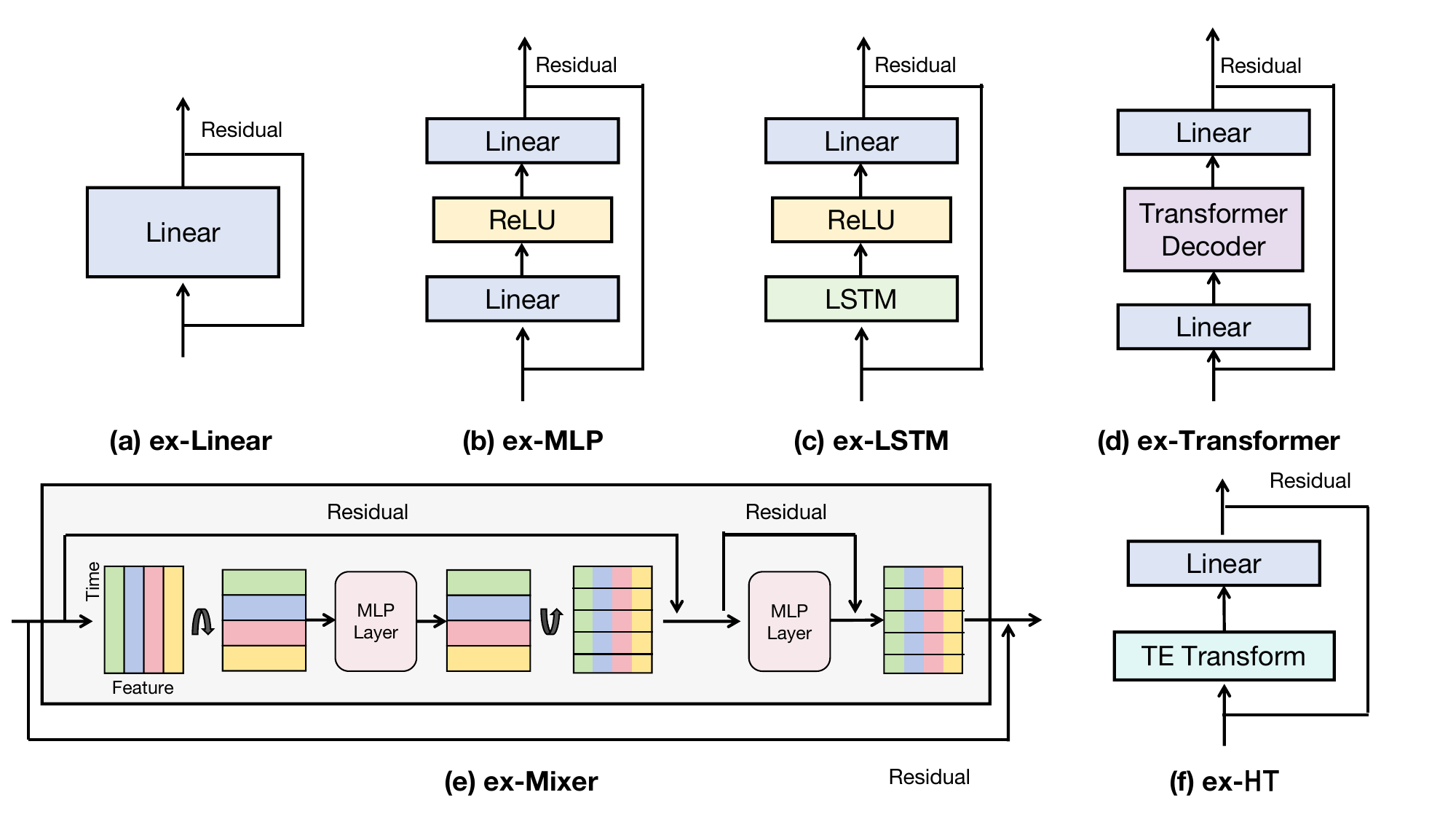}
\centering
\caption{Overview of auxiliary variables modules. (a) Single Linear Layer, (b) 2-layer MLP, (c) LSTM Layer with Linear Layers, (d) Transformer Decoder with Linear Layers, (e) Mixer Layer, (f) HT-Transform Embedding Layer designed for temperature-calender-load variables. }
\label{fig:ex_structure}
\end{figure}

Figure \ref{fig:ex_structure} summarizes the 5 neural network modules we implement, which are the commonly used Linear layer, MLP layer, LSTM layer, Transformer layer, and Mixer layer. In applications, these modules concatenate the forecasting results of the original general time series forecasting model with auxiliary variables as input. Then these modules are residually linked to the original neural network. While retaining the knowledge learned by the original time series model to the greatest extent, the design introduces the information of the auxiliary variables. Note that in implementation, we find that if we train these modules together with the forecasting models, the result will be poor. Therefore, we apply two-stage training that we will first train the forecasting models individually and then take the output of the forecasting models to train the ex-modules. In addition to the 5 neural network structures, we also introduce a specially designed embedding method to handle the situation of using temperature and calendar as auxiliary variables to forecast load. By combining different time series forecasting models and ex-modules, we can obtain 146 different methods.

\subsubsection{HT Transform: Temperature-calendar feature embedding method}
The impact of temperature on load is greatly influenced by calendar variables. Inspired by the linear regression model \cite{hong2016probabilistic} that widely acts as a benchmark in lots of load forecasting competition, we apply one-hot encoding to calendar variables and then model this coupling relationship by taking their products with temperature to the first, second, and third powers as features. The specific formula is as follows:
\begin{align}
&\hat{y}_t = \beta_0 + \beta_1 \cdot F_t+ \boldsymbol{\beta}_2^T \mathbf{M}_t + \boldsymbol{\beta}_2^T \mathbf{W}_t + \boldsymbol{\beta}_4^T \mathbf{H}_t + \boldsymbol{\beta}_5^T \mathbf{D}_t\notag\\
& + \beta_6 \cdot T_t + \beta_7 \cdot T_t^2 + \beta_8 \cdot T_t^3 + \boldsymbol{\beta}_9^T \mathbf{M}_t \cdot T_t+ \boldsymbol{\beta}_{10}^T  \mathbf{M}_t \cdot T_t^2\notag\\
&+ \boldsymbol{\beta}_{11}^T \mathbf{M}_t \cdot T_t^3+ \boldsymbol{\beta}_{12}^T  \mathbf{H}_t \cdot T_t+ \boldsymbol{\beta}_{13}^T  \mathbf{H}_t \cdot T_t^2+ \boldsymbol{\beta}_{14}^T  \mathbf{H}_t \cdot T_t^3,
\end{align}
where $\mathbf{M}_t$, $\mathbf{W}_t$, $\mathbf{H}_t$, $\mathbf{D}_t$, and $T_t$ represent the month, workday, hour, the day of the month vectors that after one-hot encoding and temperature at the corresponding time while $F_t$ is the forecasting result from the general time series forecasting model. Then $\beta_i$ is the linear coefficient, and the bold represents the vector. The reason for doing so is that the impact of temperature on load should be different in different months and hours, and such a feature embedding strategy can help the forecasting model cope with situations where the temperature and load relationship shifts under different calendar variables. To preserve such characteristics and integrate existing sequence modeling methods, we treat the information extracted by sequence modeling methods, i.e., $F_t$, as a kind of combination of the trend term and cycle term of the original time series and then concatenate them with the previously obtained calendar temperature coupling variables. Finally, a linear layer is used to map it into the final output result.

\subsubsection{Custom cost-oriented loss function}\label{cus_loss}

Unlike general time series forecasting, energy forecasting prioritizes downstream task benefits (e.g., dispatching costs) over pure accuracy. Based on \cite{zhang2022cost}, our toolkit integrates the relationship between forecasting errors and normalized real requirements $C_i \in [0,1]$ into gradient descent training via a piecewise linearization loss function $L(\epsilon)$. Specifically, we use the Forecasting Error Percentage (FEP) as our error metric: $\epsilon_i=\frac{f(x_i)-y_i}{y_i}$. 

To efficiently estimate $C$ while avoiding the discontinuities of a spline cubic function $s$, we adopt a piecewise linearization strategy. The number of segments $K$ is determined by the upper bound of the fitting error \cite{berjon2015optimal}:
\begin{equation}
\left\|s-L(\epsilon)\right\|_2 \leq \frac{\left(\int_{\epsilon_{\min}}^{\epsilon_{\max}} s^{\prime \prime}(\epsilon)^{\frac{2}{5}} d \epsilon\right)^{\frac{5}{2}}}{\sqrt{120} K^2}
\end{equation}
The interval points are distributed following \cite{de1978practical} (detailed in Section \ref{simulated}). Finally, to ensure the necessary differentiability for the loss function, we provide an option to smooth the breakpoints using a localized quadratic function, which guarantees first-derivative continuity.

Adapted from \cite{zhang2022cost}, we simulate an economic dispatch problem using a modified IEEE 30-bus system (6 generators, 3 BESS, 21 loads) \cite{hota2016analytical}. Total dispatch costs are evaluated through Day-Ahead Economic Dispatch (DAED) and Intra-day Power Balancing (IPB) optimizations, with simulation details and visualizations provided in Appendix \ref{simulated}. 

The simulation results reveal two key distinctions from the traditional MSE. First, the cost penalty is highly asymmetric: over-predictions lead to unnecessary generator over-dispatching, whereas under-predictions necessitate purchasing expensive emergency power. Second, minimizing dispatch costs does not require perfect alignment with true values, but rather satisfying specific operation bounds. Leveraging these insights, we have packaged a custom loss function derived from these simulated data, which guides the model to penalize high-cost errors and directly minimize the final dispatching cost. It is worth noting that while theoretically sound, the actual effectiveness of this custom cost-oriented loss function is highly coupled with the underlying neural network architectures. Therefore, our toolkit provides it as an optional plug-and-play module. When activated, this custom loss is combined with the traditional MSE via a weighting factor to formulate the final loss function, allowing the Agent to dynamically benchmark its utility rather than imposing it as a universal default.

\section{Energy Forecasting Knowledge Archive}
In this section, we first introduce the energy dataset we use, followed by the process of building forecasting benchmarks, how to design this knowledge into a knowledge archive that can be used by LLM agents, and finally, the visualization of the effectiveness of our designed ex-modules in the archive.
\subsection{Data Description}
To establish a comprehensive empirical foundation for the agent, we curated a large-scale benchmark comprising 21 distinct energy datasets. As summarized in Table \ref{load_dataset} and Table \ref{renew_Dataset}, these cover aggregated electrical loads, building-level micro-loads, spot market electricity prices, and renewable generation (onshore/offshore wind and photovoltaics) across various spatial resolutions. Among them, the renewable energy datasets are collected by us in the developed first-level administrative regions of China. Crucially, each dataset is meticulously paired with relevant high-quality meteorological covariates (e.g., temperature, continuous wind vectors, solar irradiance). To facilitate the discussion in the forecasting archive, we categorize the evaluated datasets into distinct domains and introduce their corresponding abbreviations. Specifically, the Aggregated Level Load (ALL) encompasses GEF12, GEF14, GEF17, PDB, Spain, Panama, and Covid19, while the Building Level Load (BLL) comprises Bull, Hog, and Cockatoo. Regarding renewable energy generation, Onshore Wind (ONW), Offshore Wind (OFFW), and Photovoltaic (PV) represent data across station, city, and regional levels. Finally, Electricity Price (EP) includes the DK1 and DK2 datasets. 

\begin{table}[htbp]\Huge
\centering
\caption{Datasets in the load and electricity price forecasting archive.}
\renewcommand{\arraystretch}{1.8}
\setlength{\tabcolsep}{4.5pt}
\resizebox{\textwidth}{!}{
\begin{tabular}{ccccccc}
\hline
   & \textbf{Dataset} & \textbf{No. of series} & \textbf{Length} & \textbf{Resolution}   & \textbf{Type} & \textbf{Auxiliary variables}     \\ \hline
1  & Covid19\cite{farrokhabadi2022day}          & 1                      & 24976           & hourly                                              & aggregated-level load   & airTemperature, Humidity, etc   \\ \hline
2  & GEF12\cite{hong2014global}             & 20                     & 21870           & hourly                                           & aggregated-level load   & airTemperature                  \\
3  & GEF14\cite{hong2016probabilistic}             & 1                      & 17520           & hourly                                       & aggregated-level load   & airTemperature                  \\
4  & GEF17\cite{hong2019global}             & 8                      & 17544           & hourly                                         & aggregated-level load   & airTemperature                  \\
5  & PDB\cite{Yeafi2021}              & 1                      & 17520           & hourly                                            & aggregated-level load   & airTemperature                  \\ \hline
6  & Spain\cite{Jhana2019}           & 1                      & 17520           & hourly                                        & aggregated-level load   & airTemperature, seaLvlPressure, etc \\
7  & Panama\cite{Shahanei2021}           & 1                      & 13033           & hourly                                        & aggregated-level load   & airTemperature, Humidity etc \\
8  & Hog\cite{Miller2020-yc}               & 24                     & 17544           & hourly                                           & building-level load     & airTemperature, wind speed, etc \\
9  & Bull\cite{Miller2020-yc}             & 41                     & 17544           & hourly                                            & building-level load     & airTemperature, wind speed, etc \\
10  & Cockatoo\cite{Miller2020-yc}         & 1                      & 17544           & hourly                                          & building-level load     & airTemperature, wind speed, etc \\ \hline
11  & DK1\footnote{\href{https://en.energinet.dk/}{https://en.energinet.dk/}}         & 1                      & 24816           & hourly                                          & electricity price     & Load, Renewable energy, etc \\ 
12  & DK2\footnote{\href{https://en.energinet.dk/}{https://en.energinet.dk/}}        & 1                      & 24816           & hourly                                          & electricity price     & Load, Renewable energy, etc \\ \hline
\end{tabular}
}\label{load_dataset}
\end{table}

\begin{table}[htbp]\Huge
\centering
\caption{Datasets in the renewable energy forecasting archive.}
\renewcommand{\arraystretch}{1.8}
\setlength{\tabcolsep}{4pt}
\resizebox{\textwidth}{!}{
\begin{tabular}{cccccccc}
\hline
  & \textbf{Data} & \textbf{No. of series} & \textbf{Type} & \textbf{Level}  & \textbf{Resolution} & \textbf{Auxiliary variables}          \\ \hline
1 & LW\_S         & 10                     & onshore wind     & station                    & 15 min              & wind speed, wind direction, relative Location \\
2 & OW\_S         & 1                      & offshore wind & station                      & 15 min              & wind speed, wind direction, relative Location \\
3 & PV\_S         & 10                     & photovoltaic            & station                    & 15 min              & irradiance, temperature, relative Location    \\ \hline
4 & LW\_C         & 16                     & onshore wind     & city                       & 15 min              & wind speed, wind direction                                    \\
5 & OW\_C         & 8                      & offshore wind & city                        & 15 min              & wind speed, wind direction                                    \\
6 & PV\_C         & 13                     & photovoltaic            & city                         & 15 min              & irradiance, temperature                                    \\ \hline
7 & LW\_R         & 5                      & onshore wind     & region                     & 15 min              & wind speed, wind direction                                    \\
8 & OW\_R         & 4                      & offshore wind & region                      & 15 min              & wind speed, wind direction                                    \\
9 & PV\_R         & 5                      & photovoltaic            & region                      & 15 min              & irradiance, temperature                                    \\ 
\hline
\end{tabular}
}\label{renew_Dataset}
\end{table}

\subsection{Forecasting Bechmarking}\label{experimental}

We execute broad benchmarking across diverse datasets representing distinct physical energy domains, comparing probability boundaries (optimized via Pinball Loss) against deterministic estimations (optimized via MSE).
\subsubsection{Task Setup \& Auxiliary Variables}
Consistent with \cite{wang2024timexer}, the look-back window and forecasting horizon are both fixed at 24 steps (applicable for day-ahead and intra-day tasks). Chronological dataset partitions rigorously follow an 0.8:0.16:0.04 ratio for training, validation, and testing. Models are trained synchronously over 50 epochs utilizing an early stopping patience of 5 constraints. For optimal grounding, specific auxiliary sequences are dynamically mapped based on target traits: scalar temperature for electrical loads; $u/v$ component vectors for wind generation; combined temperature and irradiance for photovoltaics; alongside regional loads and renewable generation distributions for spot electricity prices.
\subsubsection{Models \& Evaluation Criteria}
Multivariate forecasting leverages the 31 available architectures. Models natively incapable of extracting auxiliary data are implemented using our tailored \textit{ex-modules}, reporting overall averaged performance to ensure fair comparison. Predictive structures are predominantly evaluated utilizing Mean Absolute Error (MAE) for deterministic tracks, and cumulative Pinball Loss spanning quantiles from 0.1 to 0.9 for probabilistic scenarios. To unbox the true distribution capacities beyond absolute errors, our toolkit supports automated multidimensional metric matrices, thoroughly illustrating properties via comprehensive indicators such as Calibration Error \cite{chung2021beyond} and Winkler Score \cite{winkler1972decision}.

\subsection{Meta-Learning Based Knowledge Archive Construction}

\begin{figure}[htbp]
\centering
\includegraphics[width=\textwidth]{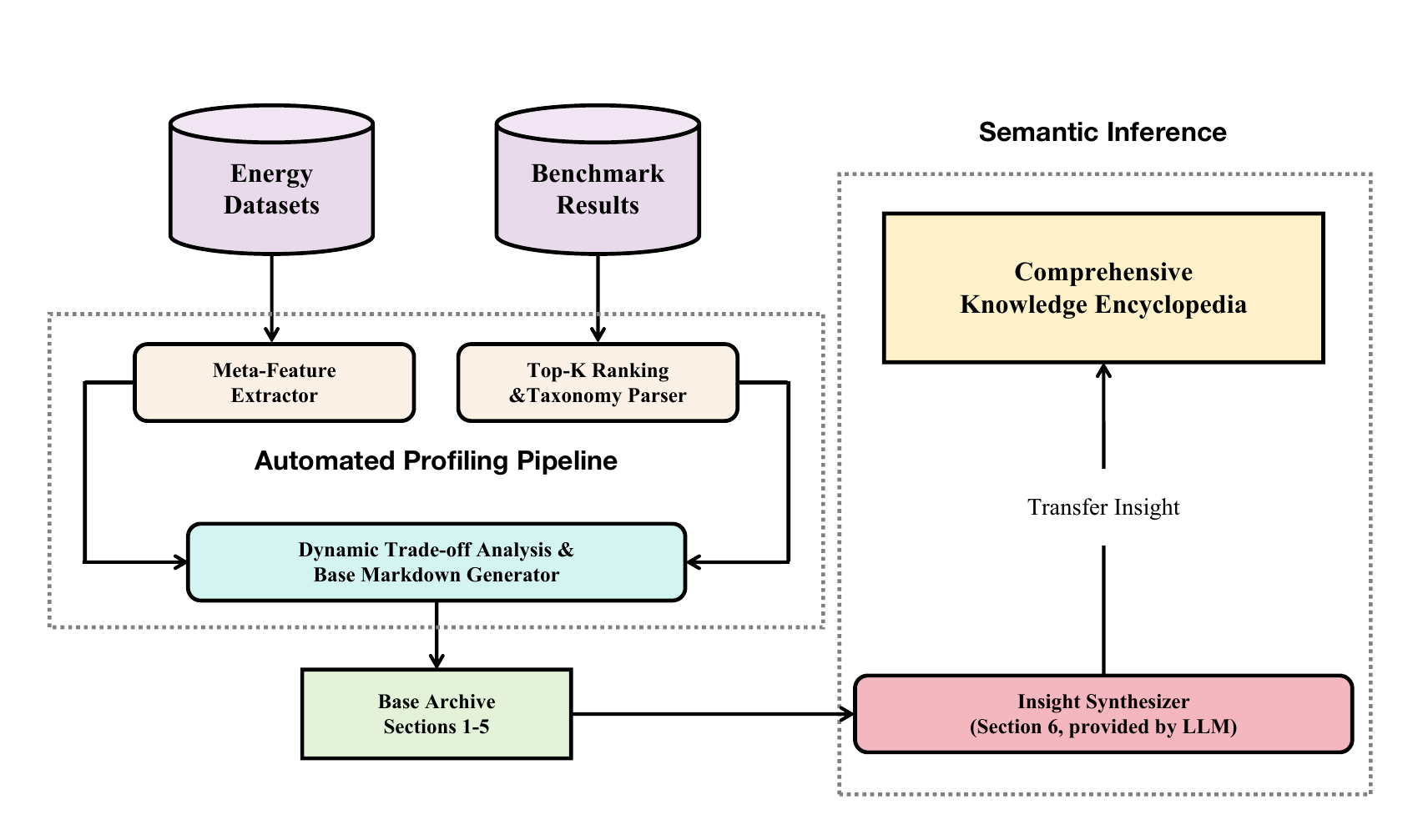}
\centering
\caption{Construction workflow of generic knowledge representation.}
\label{Forecasting_Archieve}
\end{figure}

To endow autonomous agents with robust prior knowledge and avoid the sub-optimality of heuristic manual tuning, we formulate a fully automated, data-driven meta-learning pipeline \cite{vanschoren2018meta,lemke2010meta}. Rather than relying on rigid statistical mapping, this workflow constructs a generic knowledge representation by aligning algorithmic performance with the physical profile of energy sequences.

As illustrated in Figure \ref{Forecasting_Archieve}, to structurally map the time-series physical space to model inductive biases, our pipeline generates a standardized \textbf{6-part knowledge entry} for each domain. Crucially, the first five components are deterministically computed by our underlying automated profiling script to form an objective base archive:
\begin{enumerate}
    \item \textbf{Time-Series Meta-Feature Signature:} Quantifying topological properties via spectral entropy, STL decomposition (seasonality/spikiness), and covariate cross-correlation.
    \item \textbf{Architectural Taxonomy $\&$ Covariate Directives:} Categorizing the candidate pool into decoupled combinatorial variants (e.g., $\_ex\_Linear$) versus native multivariate networks.
    \item \textbf{Evaluation Metrics Glossary:} Aligning mathematical losses (e.g., Pinball, Calibration) with operational grid intents (e.g., capacity reserve, boundary safety).
    \item \textbf{Absolute Empirical Benchmarks:} Exhaustive Top-30 leaderboard matrices across probabilistic and Top-10 for deterministic tracks.
    \item \textbf{Objective Trade-off Analysis:} Algorithmic conflict diagnosis, automatically identifying structural deviations such as general density accuracy versus outlier penalization (MAE vs. RMSE).
\end{enumerate}

Building upon this dense, mathematically rigorous foundation (Sections 1-5), an LLM is invoked as an Insight Synthesizer to deduce the final component: \textbf{6) Transferable Insights for Unseen Datasets.} Rather than merely summarizing text, the LLM critically analyzes the causal links between the initial meta-feature portraits and the empirical benchmark rankings. It autonomously synthesizes high-order semantic routing logic—for instance, linking the lack of covariate correlation directly to the absolute necessity of native dense networks (e.g., TiDE, NBEATS).

\section{Agentic Workflow}

\begin{figure}[htbp]
\centering
\includegraphics[width=\textwidth]{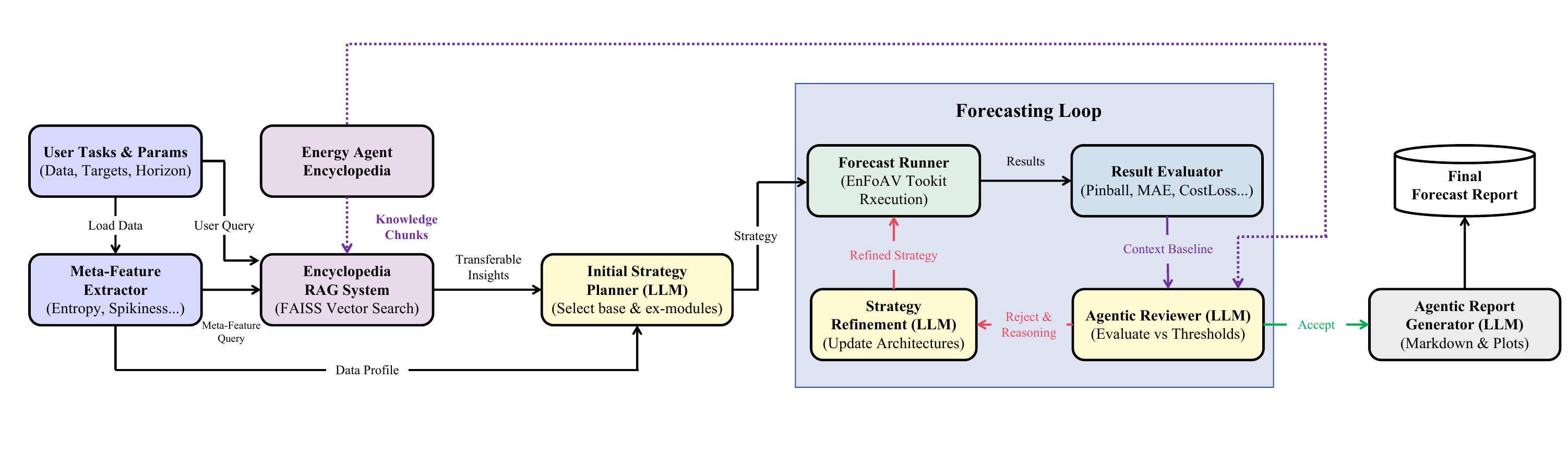}
\centering
\caption{Agentic workflow of autonomous agent.}
\label{workflow}
\end{figure}

To bridge the fundamental gap between complex machine learning architectures and practical power grid demands, we propose a closed-loop, LLM-empowered multi-agent framework. As illustrated in Figure \ref{workflow}, this framework transitions the traditional heuristic manual tuning into an autonomous reasoning paradigm, heavily inspired by recent advancements in ReAct (Reasoning and Acting) \cite{yao2022react} and Reflexion mechanisms \cite{shinn2023reflexion}.

\textbf{1) Perception and Knowledge Retrieval}: The workflow initiates by abstracting user intents to formalize strictly bounded operational requirements (e.g., target horizons, deterministic vs. probabilistic tracks, and cost-oriented constraints). Concurrently, the abstract physical signature of the raw dataset is quantified via the meta-feature extractor. To provide the agent with domain-specific grounding and minimize LLM hallucinations, we integrate a Retrieval-Augmented Generation (RAG) module \cite{lewis2020retrieval}. By executing similarity searches via FAISS in our pre-constructed Meta-Learning Encyclopedia, the system retrieves highly relevant historical mappings that translate the current data’s structural behavior into proven architectural priors.

\textbf{2) Dynamic Orchestration and Execution}: Armed with the retrieved insights and empirical data profiles, the Initial Strategy Planner operates as the cognitive orchestrator. Diverging from static heuristic searches, the planner logically synthesizes the optimal temporal backbone with customized exogenous modules (e.g., coupling a baseline sequence model with temperature-heuristic embeddings). This fully tailored strategy is subsequently delegated to the underlying Forecast Runner for mathematical optimization and boundary generation.

\textbf{3) Agentic Reflection and Self-Correction}: Instead of relying on static numeric thresholds to halt the execution, the framework employs an Agentic Reviewer acting as an expert critic. Following the Actor-Critic paradigm in LLM agents \cite{xi2025rise}, the reviewer cognitively evaluates multidimensional metrics (Pinball Loss, MAE, or IPB dispatch costs) against the retrieved baseline expectations. The framework will automatically calculate the gap between the optimal method and other methods in the current iteration. When the gap is greater than $10\%$, it will remind the reviewer to pay attention. If the convergence exhibits systemic underfitting, the reviewer formulates explicit algorithmic feedback, triggering the Strategy Refinement LLM to dynamically substitute architectures and reignite the predictive loop. 

Ultimately, upon achieving statistical convergence, the Agentic Report Generator handles the semantic synthesis. By translating dense execution logs, uncertainty boundaries, and architectural rationales into actionable, human-readable markdown reports, the workflow seamlessly aligns the complex outputs of deep temporal neural networks with the strategic decision-making needs of grid operators.
\section{Case Study}

In this section, we demonstrate the two main functions of our agentic framework. The first part is the analysis and parameter optimization of unknown data to obtain the best energy forecasting method. The second part is the comparative function, which can conduct comparative experiments and analysis based on user needs, taking the cost oriented loss function as an example. For experiment, we first introduce a closed source city level load data (to avoid unfair leakage in the training data of existing LLM), which is a 14 year load data from a developed prefecture level city in eastern China with a resolution of 1 hour. Attached is the average temperature recorded by meteorological stations within the city. In addition, all experimental settings are the same as the standard probabilistic forecasting benchmark mentioned in section \ref{experimental}. 

\subsection{Case I: Optimization of forecasting accuracy}

In order to compare the effectiveness of our \textbf{energy forecasting knowledge archive} and \textbf{energy forecasting tookit}, we have set up three sets of comparison methods. Note that all of our language models here are DeepSeek-V4.

\begin{enumerate}
    \item Group A: 31 native probabilistic forecasting model mentioned in Table \ref{fig:Forecasting_models}.
    \item Group B: Native LLM with our energy forecasting toolkit.
    \item Group C: LLM with RAG (energy forecasting knowledge archive) and our energy forecasting toolkit.
\end{enumerate}

\begin{figure}[htbp]
\centering
\includegraphics[width=\textwidth]{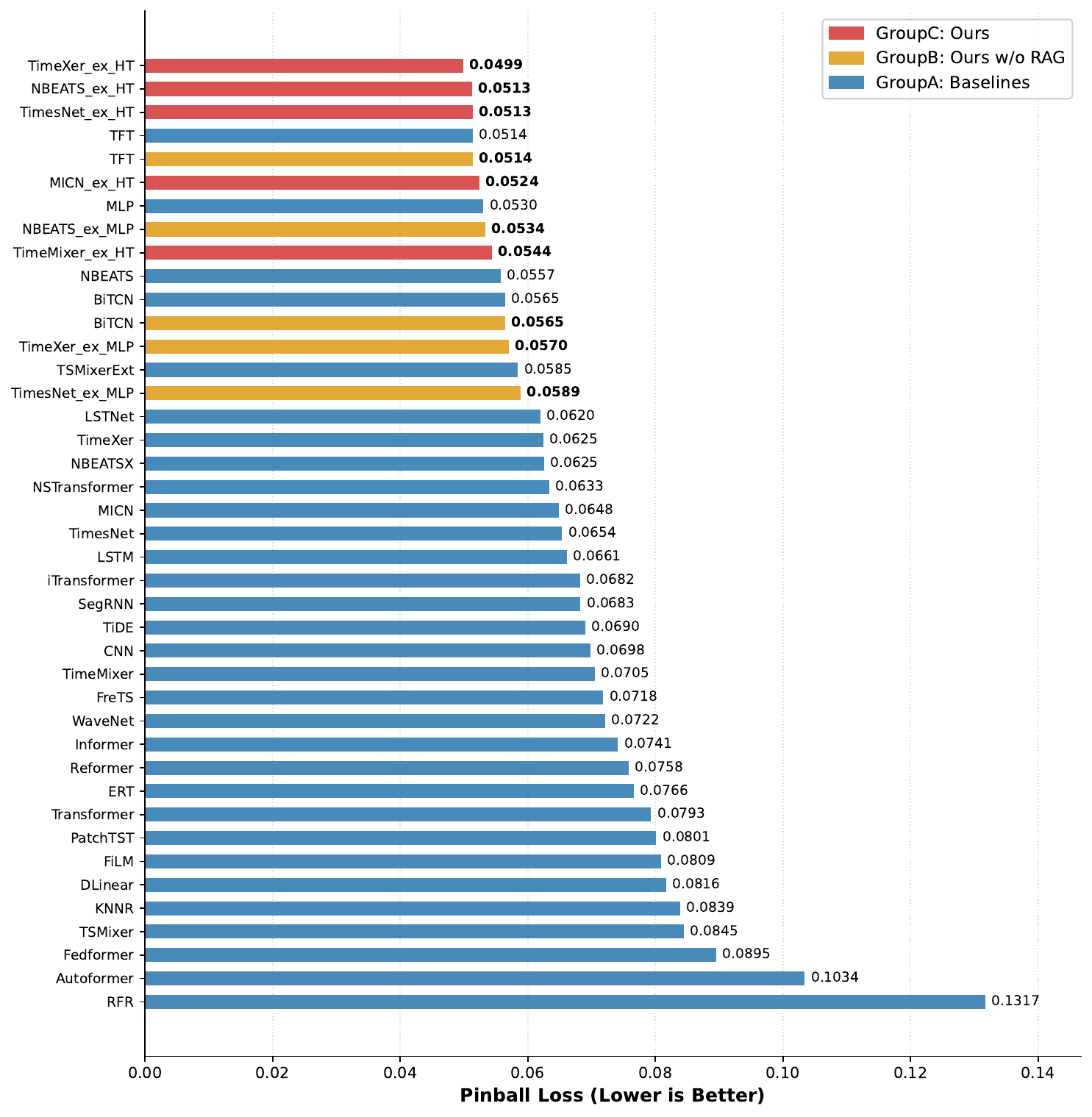}
\centering
\caption{Performance comparison between the original forecasting model and the forecasting model with our workflow.}
\label{Agentic_performance}
\end{figure}

Prompt \ref{prompt_forecasting} shows the user's requirement description. Figure \ref{Agentic_performance} reports the probabilitic forecasting performance of the original Vanila model and the model within our workflow. Overall, Group C holds a clear absolute advantage: its models occupy the top three ranks, and its lowest-ranking model still places 9th. Furthermore, the comparison between Group C and Group A highlights the inherent advantage of our ex-module in handling auxiliary variables. Notably, multiple models equipped with this module outperformed the optimal baseline, TFT, despite TFT natively supporting auxiliary features. Additionally, the contrast between Group C and Group B demonstrates the critical impact of our energy forecasting knowledge archive. Even with equivalent tool access, the RAG-empowered framework yielded significantly better forecasting performance than depending on the baseline deductive capabilities of a vanilla LLM. Table \ref{consumption_comparision} illustrates the time efficiency and resource consumption across the three groups. As expected, Group C incurs slightly higher token usage than Group B due to the retrieval-augmented generation (RAG) processes. However, at the current operational scale, this additional token overhead is practically negligible. More importantly, through LLM's targeted planning, both Group B and Group C have shortened their execution time. Especially for Group C, which is almost one-third of Group A. Group B lacks our knowledge base, which leads to LLM not having enough confidence. Therefore, it performs multiple iterations, ultimately resulting in longer time consumption.

\begin{promptbox}[label=prompt_forecasting]{Describe the forecasting requirement}
\sffamily\small 
My data path is \texttt{./data/test/load\_with\_weather.pkl}. The target column is load, and all remaining columns (temperature) are auxiliary features. Please refer to the encyclopedia, find out the structure that can deal with the temperature well. I need to make probabilistic predictions. Note that choose the best 5 model to achieve the best pinball loss.
\end{promptbox}

\begin{table}[htbp]
\centering
\caption{Method consumption comparison.} 
\label{consumption_comparision}
\renewcommand{\arraystretch}{1.3} 
\begin{tabular}{lccc}
\toprule
\textbf{Method} & \textbf{Time (s)} & \textbf{Tokens Used} & \textbf{Cost (\$)} \\ \midrule
Baseline (Vanilla) & 15287 & -- & -- \\
Ours w/o RAG & 11148 & 15624 & 0.0036 \\
Ours (with RAG) & 5599 & 18394 & 0.0037 \\ \bottomrule
\end{tabular}
\end{table}

\subsection{Case II: Autonomous Comparative A/B Testing}

To demonstrate the framework's capability as an automated testbed, we instruct the virtual analyst to evaluate the empirical economic benefits of the custom cost-oriented loss function across different neural architectures.

\begin{promptbox}[label=prompt_test1]{Describe the requirement for test different models}
\sffamily\small 
My data path is \texttt{./data/test/load\_with\_weather.pkl}. The target is load, and all remaining columns are covariates(temperature).  This is an aggregated load. I want to run a comparative A/B test: Phase 1 should be a deterministic forecast with default loss, and Phase 2 should be a deterministic forecast with cost-oriented loss. Please evaluate on 3 models.
\end{promptbox}

Upon receiving this concise prompt, the agent autonomously orchestrates the entire experimental workflow: it instantiates the data pipelines, manages the rigorous two-phase training process (default MSE/MAE vs. custom Cost-Oriented Loss), and evaluates the models across a comprehensive set of statistical and business-centric metrics. As efficiently synthesized in Table \ref{tab:ch6:ab_test}, the agent's autonomous diagnosis reveals a critical insight: the effectiveness of the cost-oriented loss function is highly architecture-dependent rather than universally beneficial. The agent autonomously identifies that architectures capable of capturing complex temporal variations, such as \textbf{PTimesNet\_ex\_Linear} and \textbf{PTimeMixer\_ex\_Linear}, possess a strong symbiotic relationship with the cost-oriented loss. \textbf{PTimesNet\_ex\_Linear} emerges as the optimal choice, achieving an impressive absolute reduction of 0.004 in CostLoss. More importantly, the agent's multi-metric tracking reveals that this business-cost improvement does not come at the expense of forecasting accuracy; rather, traditional statistical metrics such as MAE (-0.004) and MASE (-0.024) improve simultaneously. This case profoundly illustrates the framework's core value: it goes beyond simple code generation to act as an automated algorithmic scientist. By seamlessly discovering algorithmic suitability—or unsuitability—across multiple dimensions, it spares researchers from manually scripting tedious A/B testing loops and designing multi-metric evaluation pipelines. The comprehensive markdown analysis report, generated entirely by the agent with detailed visualizations and recommendations, is provided in the supplementary materials.

\begin{table}[htbp] 
\centering

\caption{A/B test on cost-oriented loss function extract from our LLM generated report.}
\label{tab:ch6:ab_test}
\renewcommand{\arraystretch}{1.0} 

\resizebox{\textwidth}{!}{ 
\begin{tabular}{@{} l l c c c @{}}
\toprule
\textbf{Model} & \textbf{Metric} & \textbf{Phase 1 (Default Loss)} & \textbf{Phase 2 (Cost-Oriented Loss)} & \textbf{$\Delta$ (Phase2 -- Phase1)} \\
\midrule
\textbf{PNBEATS\_ex\_Linear}   & MAE      & 0.138 & 0.137 & -0.001 \\
                               & RMSE     & 0.211 & 0.223 & +0.012 \\
                               & MASE     & 0.785 & 0.780 & -0.005 \\
                               & CostLoss & 0.093 & 0.095 & +0.002 \\
\midrule
\textbf{PTimesNet\_ex\_Linear} & MAE      & 0.140 & 0.135 & -0.004 \\
                               & RMSE     & 0.219 & 0.218 & -0.001 \\
                               & MASE     & 0.793 & 0.769 & -0.024 \\
                               & CostLoss & 0.098 & 0.094 & -0.004 \\
\midrule
\textbf{PTimeMixer\_ex\_Linear}& MAE      & 0.148 & 0.146 & -0.002 \\
                               & RMSE     & 0.232 & 0.229 & -0.003 \\
                               & MASE     & 0.842 & 0.830 & -0.012 \\
                               & CostLoss & 0.109 & 0.106 & -0.003 \\
\bottomrule
\end{tabular}
} 
\end{table}

\section{CONCLUSIONS}
In this paper, we propose an innovative LLM-empowered autonomous forecasting framework that bridges the fundamental gap between advanced machine learning algorithms and practical power grid operations. Rather than relying on heuristic manual tuning, our virtual analyst agent autonomously orchestrates optimal forecasting pipelines, spanning from data preprocessing to profound analytical report generation. To provide a rigorous operational foundation for the agent, we developed a highly accessible, energy-centric toolkit integrating 31 distinct architectures, domain-specific feature embeddings, and probabilistic capabilities. Crucially, the framework acts as an autonomous testbed, seamlessly conducting A/B testing to verify the empirical suitability of specific algorithmic plugins—such as our customized cost-oriented loss function—against highly variable grid data. Supported by a comprehensive zero-shot knowledge archive built upon 21 diverse energy datasets and newly released localized renewable datasets, this open-source initiative significantly lowers the deployment barrier of state-of-the-art AI technologies in modern smart grids, serving as an invaluable asset for both researchers and power industry operators.

\clearpage
\appendix
\section{Dataset Description}\label{Dataset_description}
\subsection{Electrical load dataset and electricity price dataset}
As shown in Table \ref{load_dataset}, we collect multiple load datasets at different levels as well as electricity price datasets and organize them into a user-friendly `.pkl' format and users will be able to obtain them through the URL we provide after publication. All data in this section are available under CC BY 4.0 except the PDB and Spain. These two datasets are applicable to CC0 1.0. Now we will introduce their sources and detailed information one by one.
\subsubsection{GEF12}
The GEF12 dataset is sourced from the Global Energy Forecasting Competition 2012 \cite{hong2014global}. This competition has multiple tracks, and we have compiled the dataset provided by the load forecasting tracks as one of our benchmark datasets. In this dataset, there are a total of 20 aggregated-level load series data and 11 temperature series. It is worth noting that the one-to-one correspondence between these temperature data and load data has not been clearly defined. Each time series covers load data with a resolution of 1 hour from 0:00 on January 1, 2004, to 5:00 on June 30, 2008. Because this dataset is used for competitions and the integrity of the data is relatively good, we did not preprocess the data (such as filling in missing values).

\begin{figure}[htb]
\centering
\includegraphics[width=\textwidth]{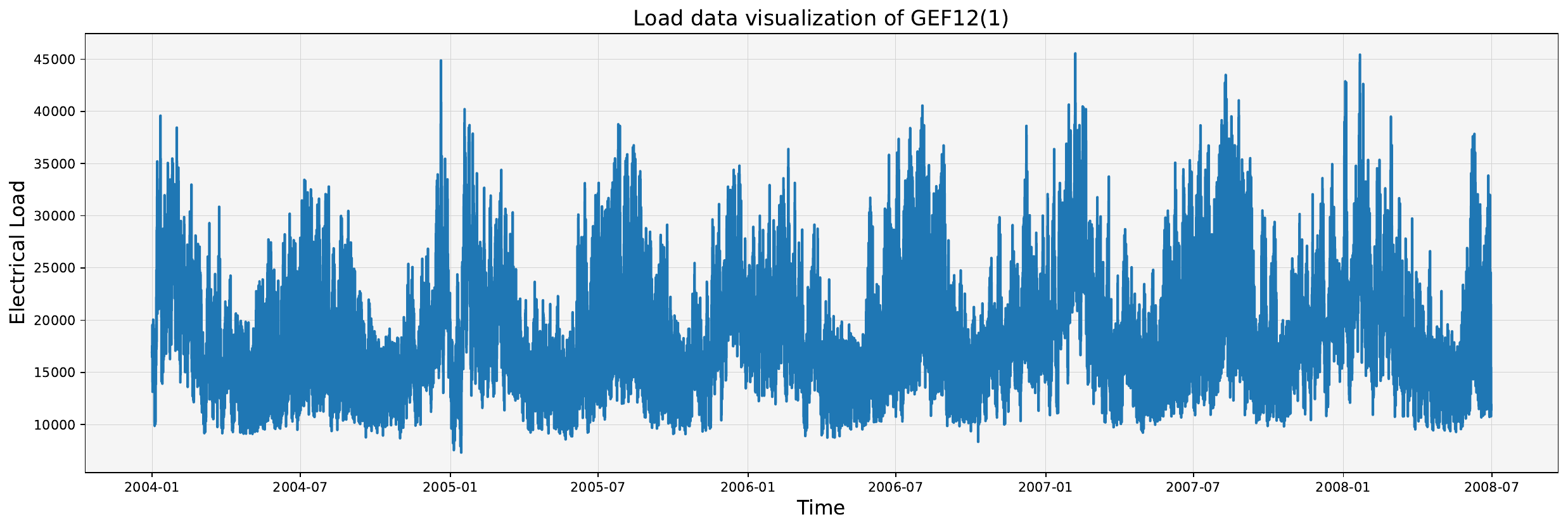}
\centering
\caption{Visualization of the first load series in the GEF12 dataset.}
\label{GEF12_1}
\end{figure}

\begin{figure}[htb]
\centering
\includegraphics[width=\textwidth]{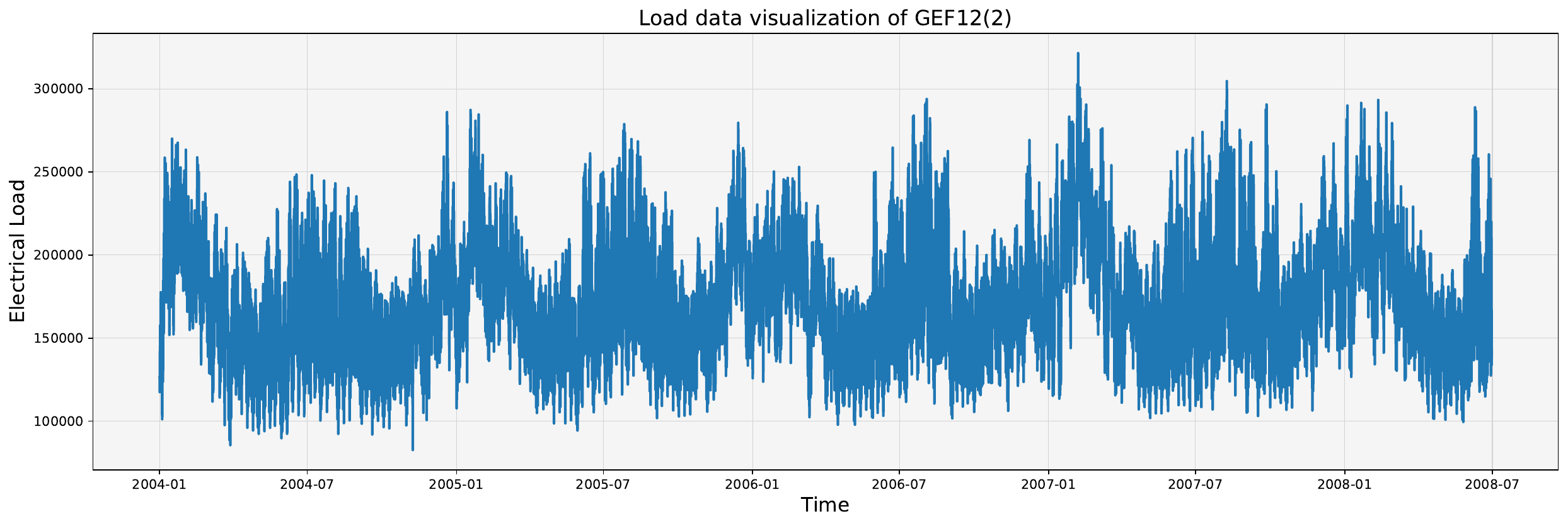}
\centering
\caption{Visualization of the second load series in the GEF12 dataset.}
\label{GEF12_2}
\end{figure}

Figure \ref{GEF12_1} and Figure \ref{GEF12_2} visualize the partial load sequence in the GEF12 dataset. From it, we can see that these load series have obvious periodicity and seasonality. And this is an important feature of aggregated-level load.

\subsubsection{GEF14}
\begin{figure}[htb]
\centering
\includegraphics[width=\textwidth]{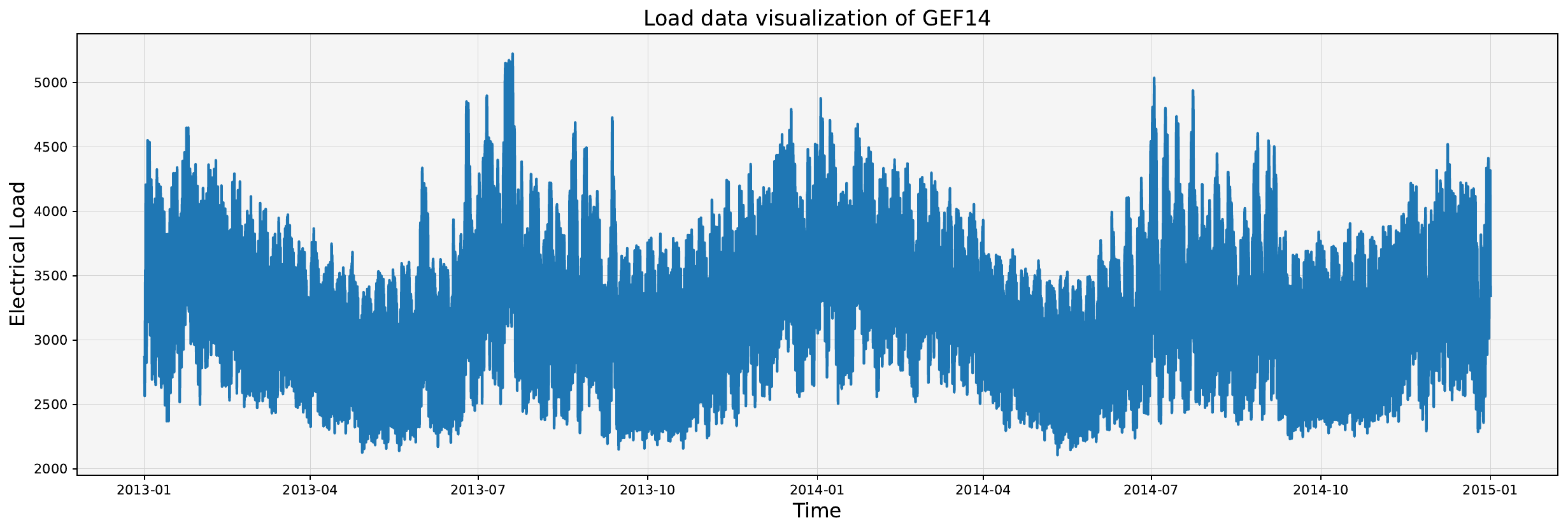}
\centering
\caption{Visualization of the load series in the GEF14 dataset.}
\label{GEF14}
\end{figure}

The GEF14 dataset is from the Global Energy Forecasting Competition 2014 \cite{hong2016probabilistic}. This competition also has multiple tracks, and we focus on load forecasting tracks. The competition provides load data spanning up to 8 years from 2006 to 2014. Unlike the 2012 competition, We truncate the load data and only use the data from 2013 and 2014 for testing. On the one hand, it is because the impact of load data from many years ago on the current forecast is very small, and on the other hand, it is because most of the load data we collect is about 2 years in length. For relative consistency, we only took the last two years to construct our load forecast archive. Our adjusted load data covers load data with a resolution of 1 hour from 1:00 on January 1, 2013, to 0:00 on January 1, 2015. Figure \ref{GEF14} shows the adjusted load data, similar to the data in GEF12, which is also aggregated level data. Therefore, the data in GEF14 also shows obvious periodicity and seasonality.

\subsubsection{GEF17}

\begin{figure}[htb]
\centering
\includegraphics[width=\textwidth]{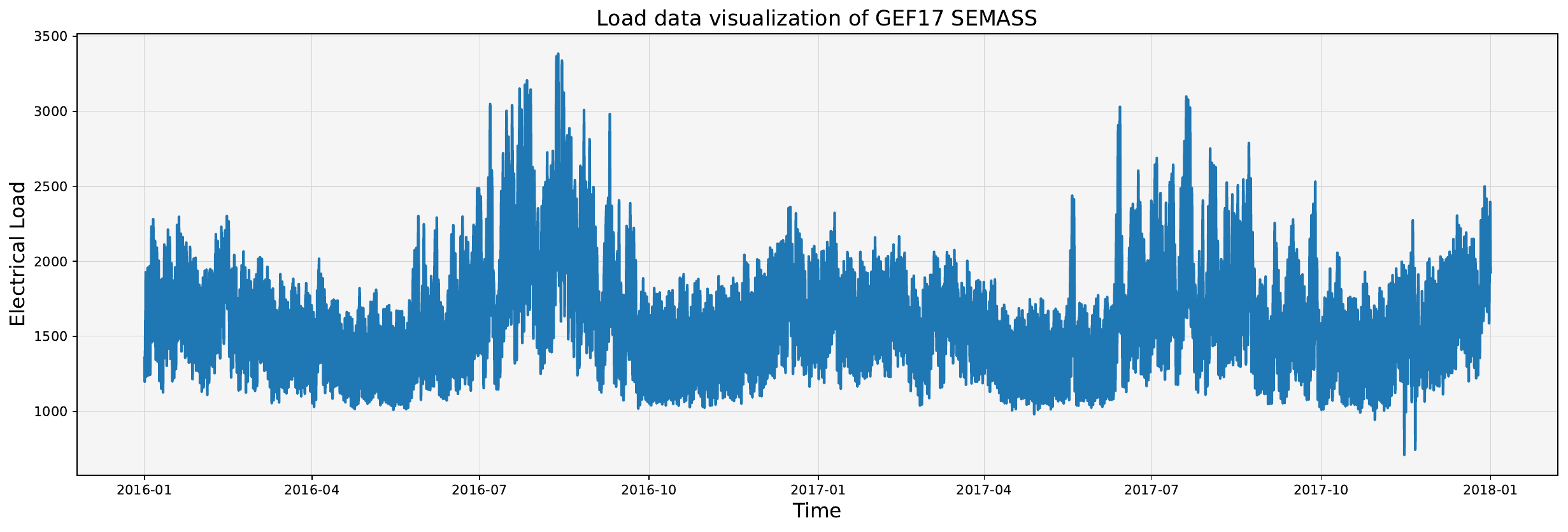}
\centering
\caption{Visualization of the load series in the GEF17 SEMASS dataset.}
\label{GEF17_1}
\end{figure}

\begin{figure}[htb]
\centering
\includegraphics[width=\textwidth]{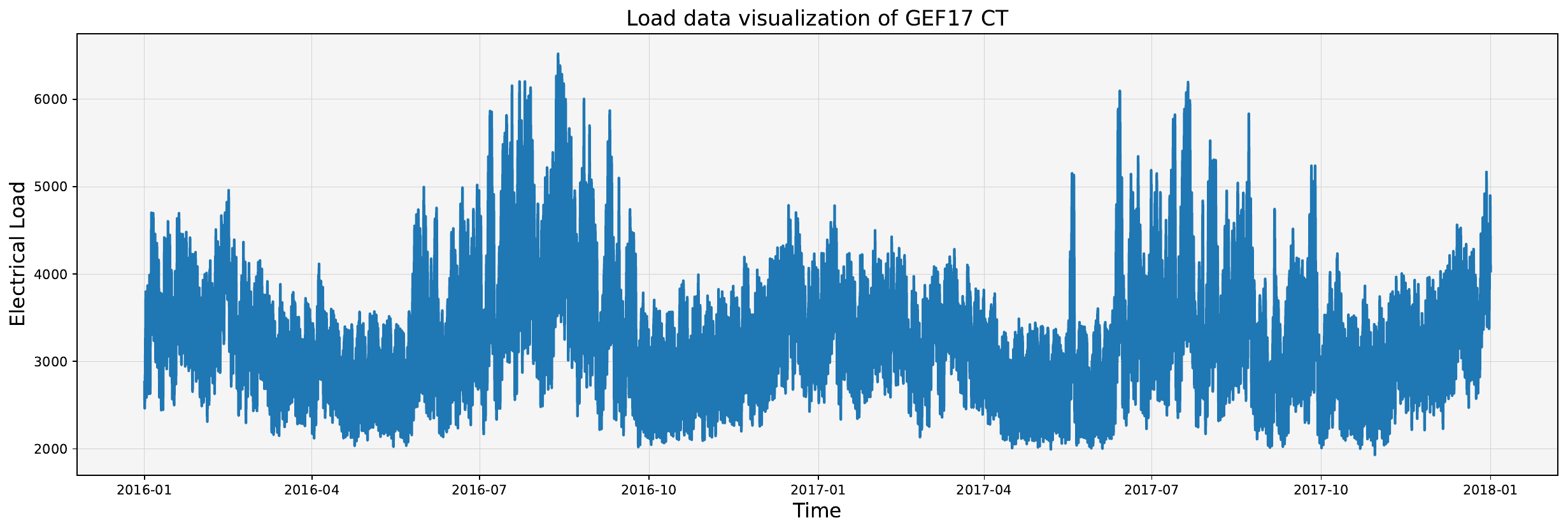}
\centering
\caption{Visualization of the load series in the GEF17 CT dataset.}
\label{GEF17_2}
\end{figure}

The GEF17 dataset is from the Global Energy Forecasting Competition 2017 \cite{hong2019global}. Similar to GEF12, this dataset also provides multiple aggregated-level load data. However, the difference is that it clarifies the corresponding relationship between the temperature series and load series, providing a one-to-one temperature series corresponding to the load series. In terms of period, it provides load data from 2013 to 2017. For the reasons mentioned above, we have intercepted the load data and only used the load data from the past two years (i.e. 2016 and 2017). Finally, we used 8 aggregated-level load data from 2016 to 2017 and their corresponding temperature data to construct our load forecasting archive. Figure \ref{GEF17_1} and Figure \ref{GEF17_2} visualize some data in the GEF17 dataset, similarly, it also showcases the common characteristics of aggregated-level load data, namely periodicity and seasonality.

\subsubsection{Covid19}

\begin{figure}[htb]
\centering
\includegraphics[width=\textwidth]{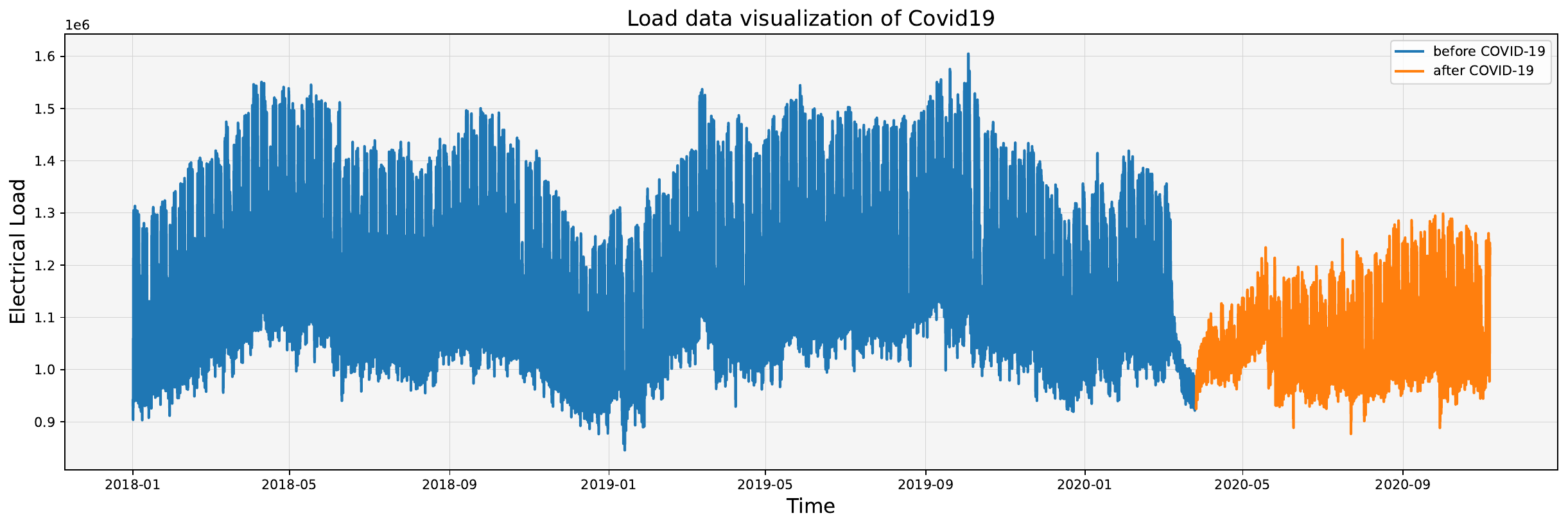}
\centering
\caption{Visualization of the load series in the Covid-19 dataset.}
\label{Covid19}
\end{figure}

The Covid-19 dataset is from the Day-Ahead Electricity Demand Forecasting Competition: Post-COVID Paradigm \cite{farrokhabadi2022day}. This dataset covers the load data from 0:00 on January 1, 2018, to 15:00 on November 6, 2020. In addition to load data and temperature, this dataset also provides other meteorological factors such as humidity and wind speed. To maintain the consistency of the forecasting archives, we did not consider such factors. Unlike the datasets mentioned above, this dataset focuses on the impact of COVID-19 on the power system. Figure \ref{Covid19} shows the load data in the Covid19 dataset. The blue part indicates that the power system has not yet been impacted by COVID-19, similar to other aggregated level load data, showing periodicity. The orange section displays the load data after COVID-19. It can be seen that it is different from the blue part. The absolute value of the load rapidly decreases during the period being impacted, and then recovers smoothly after a period of time. However, compared to the same period when it was not impacted, the load value has decreased. This transformation poses a challenge to the robustness of forecasting models.

\subsubsection{PDB}
\begin{figure}[htb]
\centering
\includegraphics[width=\textwidth]{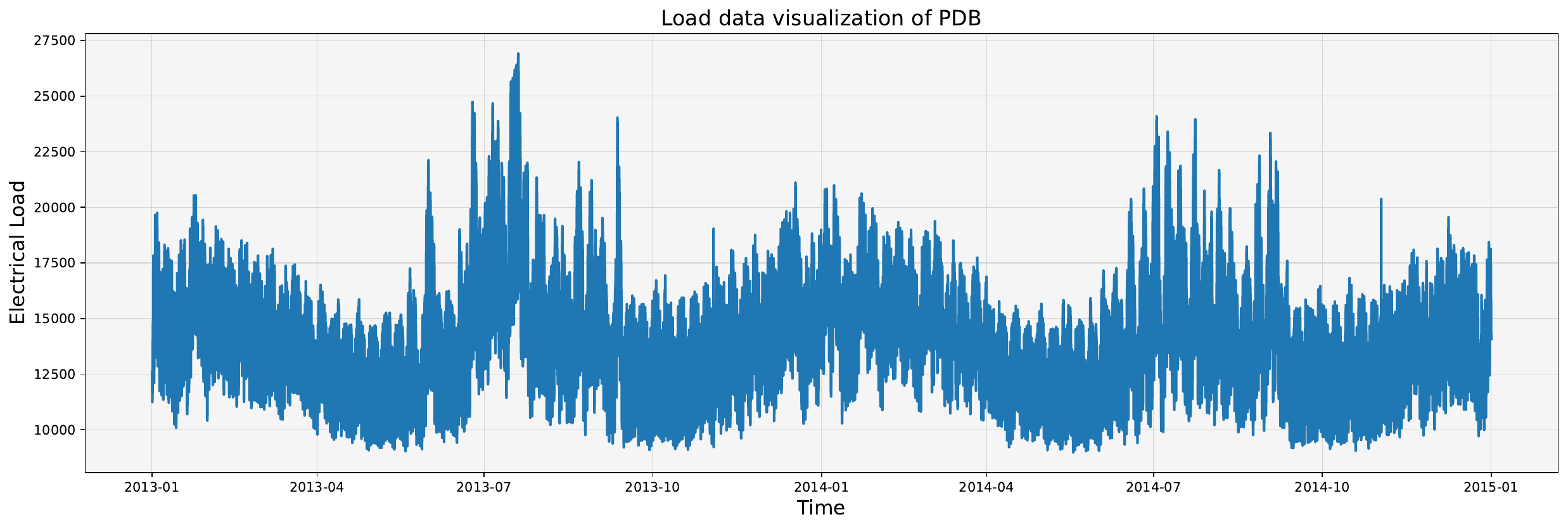}
\centering
\caption{Visualization of the load series in the Spanish dataset.}
\label{PDB}
\end{figure}

The PDB dataset is a public dataset from the Kaggle data competition platform \cite{Yeafi2021}. It covers load and temperature data from 1:00 on January 1, 2013, to 0:00 on January 1, 2015. Due to its moderate length, we did not intercept it. Figure \ref{PDB} shows its load data visualization results.

\subsubsection{Spain}

\begin{figure}[htb]
\centering
\includegraphics[width=\textwidth]{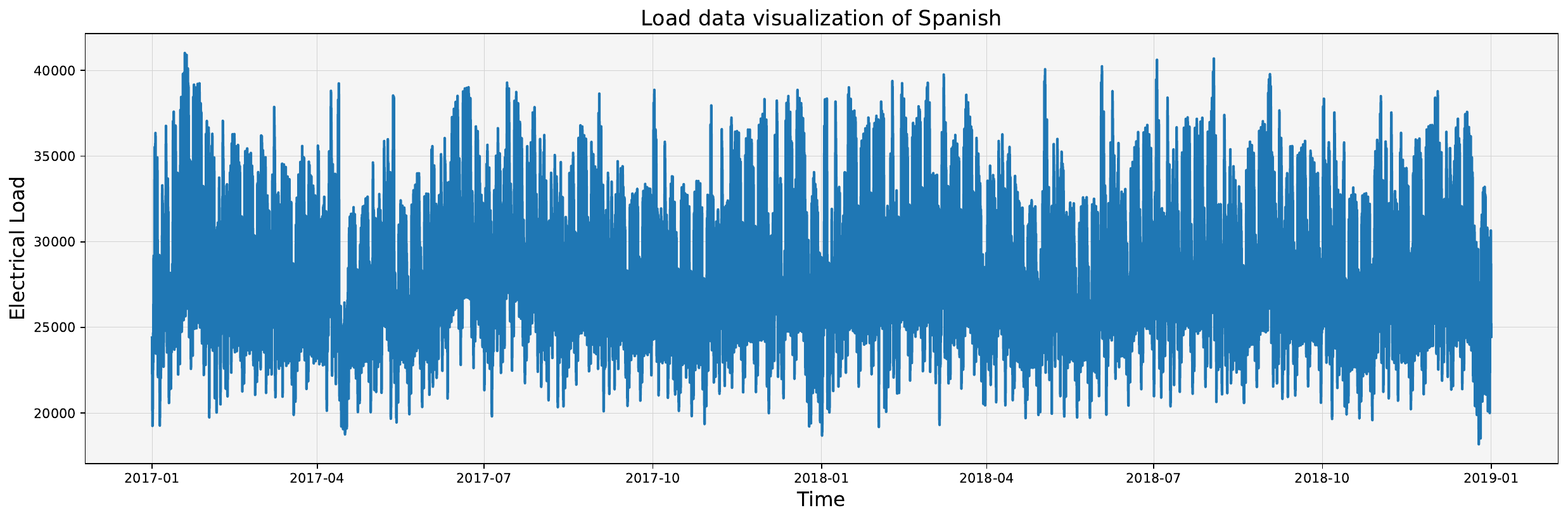}
\centering
\caption{Visualization of the load series in the PDB dataset.}
\label{Spain}
\end{figure}

The Spanish dataset is also a public dataset from the Kaggle data competition platform \cite{Jhana2019}. It provides nationwide load data for Spain from 0:00 on January 1, 2017, to 23:00 on December 31, 2018. At the same time, it also provides meteorological data (such as temperature and wind speed) corresponding to the five major cities in Spain. This situation is similar to GEF12. Note that the load data of this dataset is partially missing (with a missing rate of 0.1$\%$). Because of the low missing rate, we used a simple Linear interpolation method to fill the data. Figure \ref{Spain} shows the corresponding load visualization results. Compared to other aggregated level loads, the periodicity and seasonality of the Spanish national load have become relatively less pronounced.

\subsubsection{Panama}
\begin{figure}[htb]
\centering
\includegraphics[width=\textwidth]{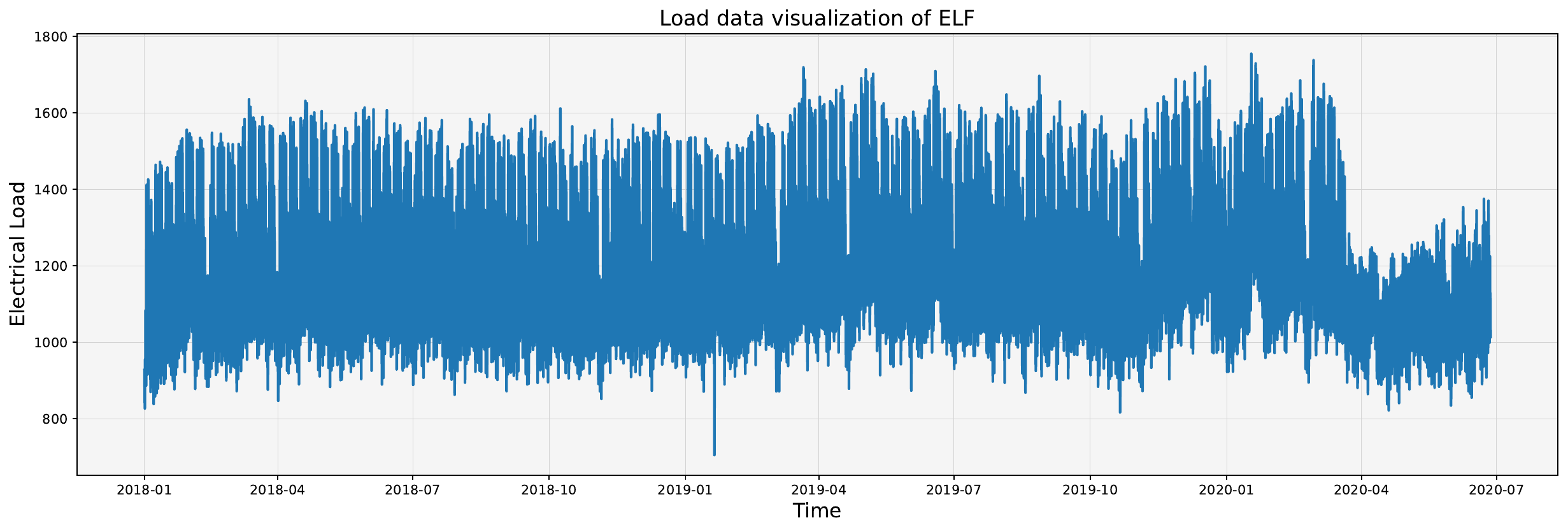}
\centering
\caption{Visualization of the load series in the Panama dataset.}
\label{ELF}
\end{figure}

The Panama dataset comes from the Kaggle data platform \cite{Shahanei2021}. It is worth noting that the platform provides temperature data for multiple cities in Panama, as well as other meteorological data such as wind speed and humidity. Like datasets such as Spain, the geographical scale of load data and meteorological data do not clearly correspond. Figure \ref{ELF} shows the load data of this dataset. Note that in our experiments, we only used data from 2019 and later. Similar to the COVID-19 dataset, this dataset also shows the impact of COVID-19 on the power system (see data after April 2020).

\subsubsection{Hog}

The Hog dataset comes from The Building Data Genome 2 (BDG2) Data-Set \cite{Miller2020-yc}. BDG2 is an open dataset that includes building-level data collected from 3053 electricity meters, which covers 1636 buildings. From the perspective of the area where the building is located, it includes the load, cooling, and heating data of buildings in multiple areas such as Hog and Bull. From a period perspective, it covers data from 2016 and 2017. In addition, BDG2 also classifies buildings, including buildings for educational purposes, offices, and so on. Based on the characteristics of this dataset, we divide it by region, and the Hog dataset is composed of relevant load data from buildings in the Hog region in the BDG2 dataset. The data in this dataset is all building-level data that we often find situations such as missing values and outliers in data at this level\cite{jeong2021missing}. Therefore, we first use the functions provided by the package to check for outliers. Specifically, we first calculate the lower quartile (Q1) and the upper quartile (Q3) and then calculate the quartile interval (IQR), that is, $IQR=Q3 - Q1$. Here, the outlier is defined as the point that is lower than $ Q1- q \times IQR$ or higher than $Q3+q\times IQR$. The outlier factor $q$ here is set to 1.5. We set the detected outlier as the missing value, and discard the sequence with a missing rate of more than 10$\%$. For sequences with a missing rate of less than $10\%$, we interpolate them (using linear, polynomial, etc.). Finally, we obtained 24 available load sequences and their corresponding temperature sequences for the Hog region. 

\begin{figure}[htb]
\centering
\includegraphics[width=\textwidth]{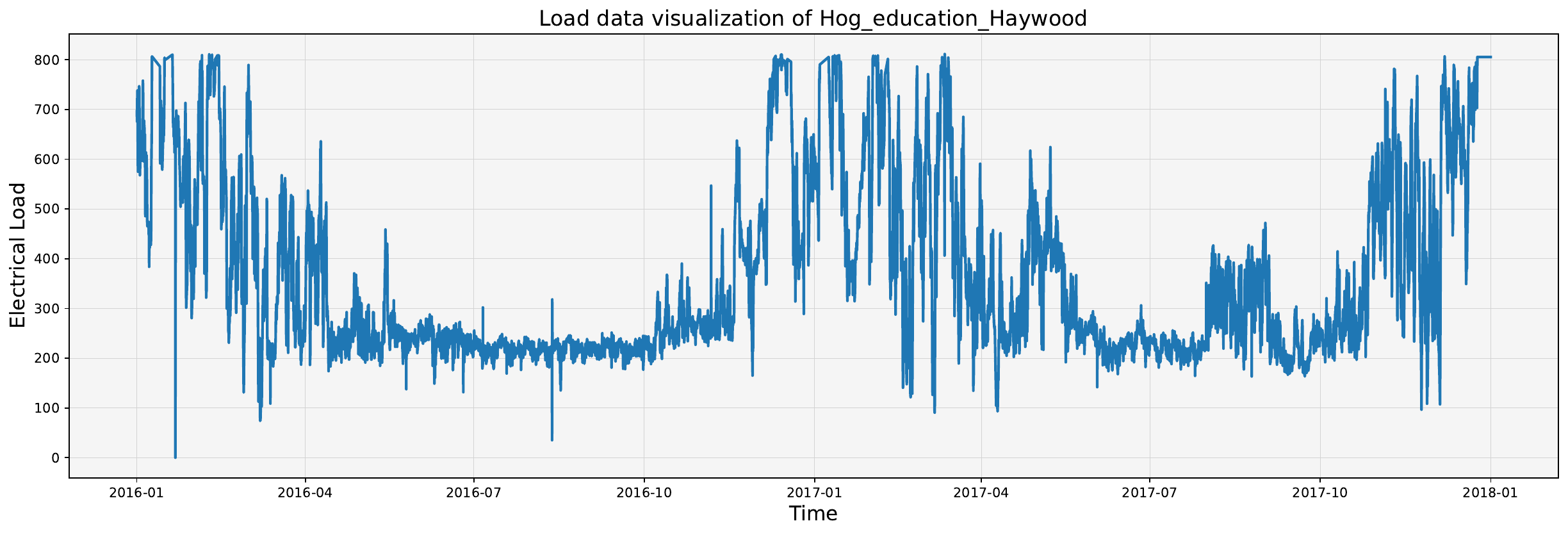}
\centering
\caption{Visualization of the Hog education Haywood in the Hog dataset.}
\label{Haywood}
\end{figure}

\begin{figure}[htb]
\centering
\includegraphics[width=\textwidth]{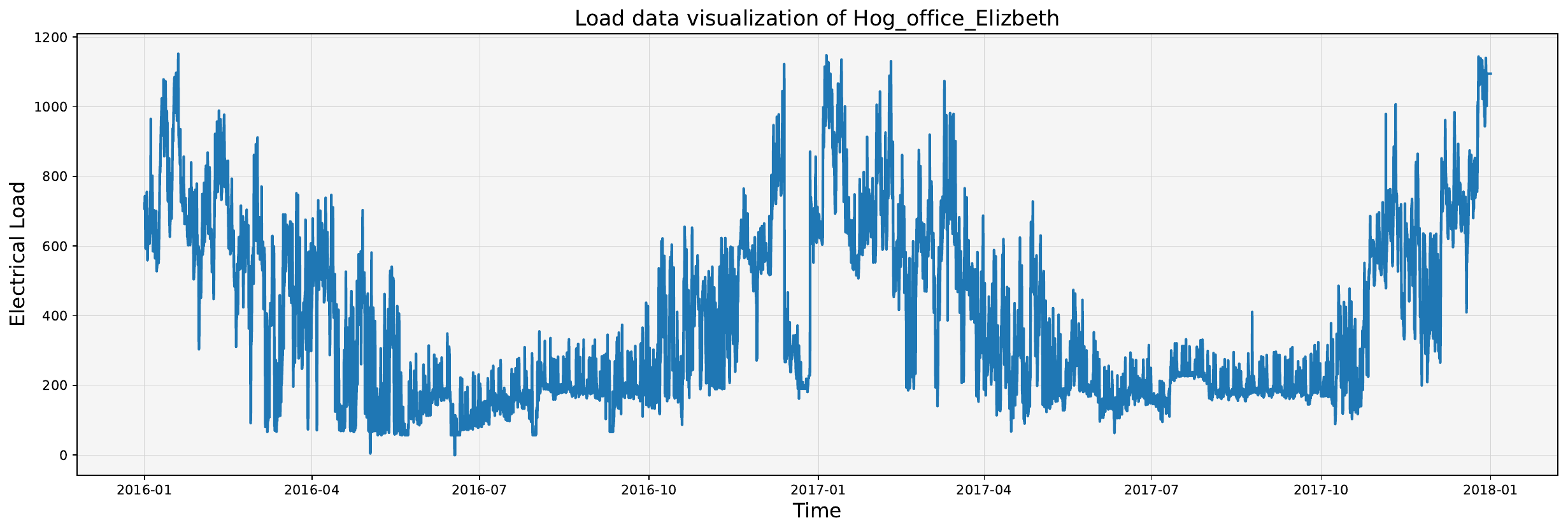}
\centering
\caption{Visualization of the Hog office Elizbeth series in the Hog dataset.}
\label{Elizbeth}
\end{figure}

Figure \ref{Haywood} and Figure \ref{Elizbeth} show two load sequences in the Hog dataset. They belong to educational facilities and offices respectively. It can be seen that compared to aggregate-level datasets, building-level datasets exhibit greater uncertainty. The similarity of data for the same period in different years is also significantly lower than those aggregate-level ones. Although the data is only two years old, the building dataset also exhibits significant seasonality. Specifically, the load during summer and autumn is relatively high, while the load during winter and spring is relatively low. In addition, despite the different properties of buildings, they still maintain a relatively similar seasonality.

\subsubsection{Bull}
\begin{figure}[htb]
\centering
\includegraphics[width=\textwidth]{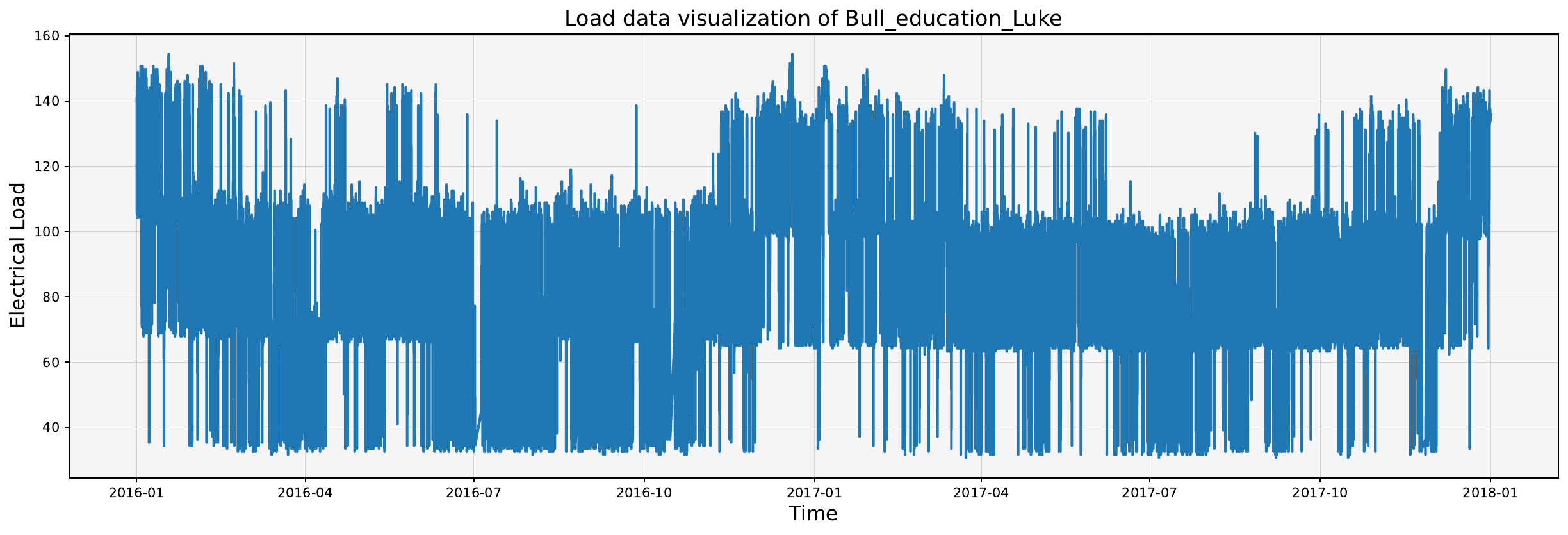}
\centering
\caption{Visualization of the Bull education Luke series in the Bull dataset.}
\label{Luke}
\end{figure}

\begin{figure}[htb]
\centering
\includegraphics[width=\textwidth]{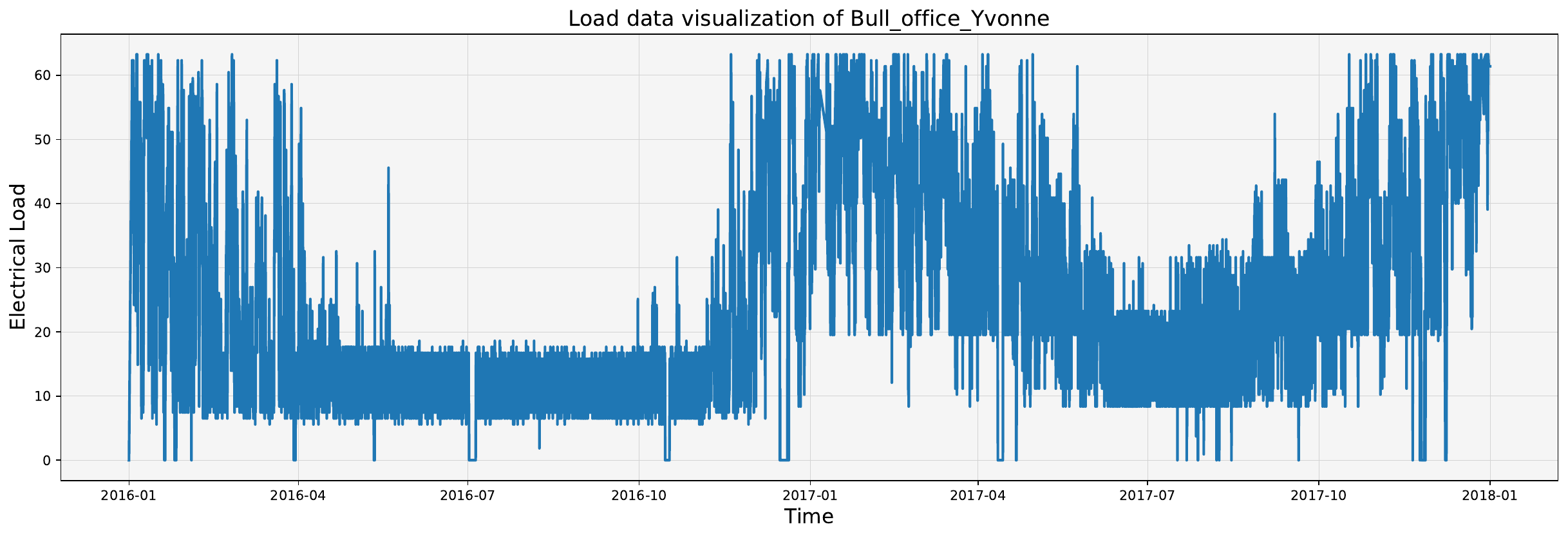}
\centering
\caption{Visualization of the Bull office Yvonne series in the Bull dataset.}
\label{Yvonne}
\end{figure}

Similar to the Hog dataset, the Bull dataset also comes from the BDG2 dataset. Similarly, we screen and preprocess the building load data in the Bull area, resulting in 41 available sequences covering multiple building properties. Figure \ref{Luke} and Figure \ref{Yvonne} show the load data of two representative building types in the Bull area. Similar to other building-level load data, the manifestation of periodicity is not obvious. Meanwhile, sudden changes also occur from time to time, posing challenges for the forecasting model to accurately model and forecast. 

\subsubsection{Cockatoo}
\begin{figure}[htb]
\centering
\includegraphics[width=\textwidth]{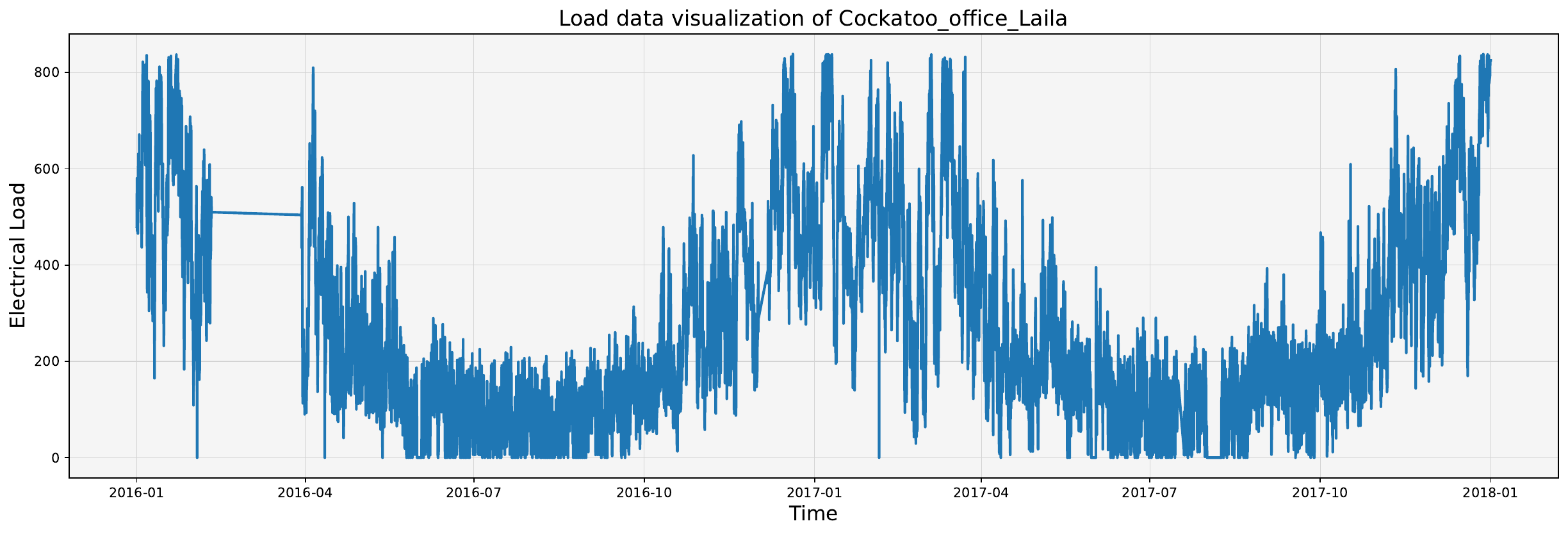}
\centering
\caption{Visualization of the Cockatoo office Laila series in the Cockatoo dataset.}
\label{Laila}
\end{figure}

Cockatoo is also from the BDG2 dataset. However, after our screening, only one load sequence met our requirements, which is “Cockatoo Office Laila”. Figure \ref{Laila} shows the load characteristics of this building. It is worth noting that during the period from February to April 2016, the load data appeared relatively stable. This may be caused by the fault of the measuring meters, by human error in the reading, or it may be the real situation. These errors typically occur in building-level load data and it is difficult to avoid this situation through data cleansing unless we directly discard the relevant data. Here, we retain this data to evaluate its impact on the final forecasting performance.

\subsubsection{DK1 and DK2}
\begin{figure}[htb]
\centering
\includegraphics[width=\textwidth]{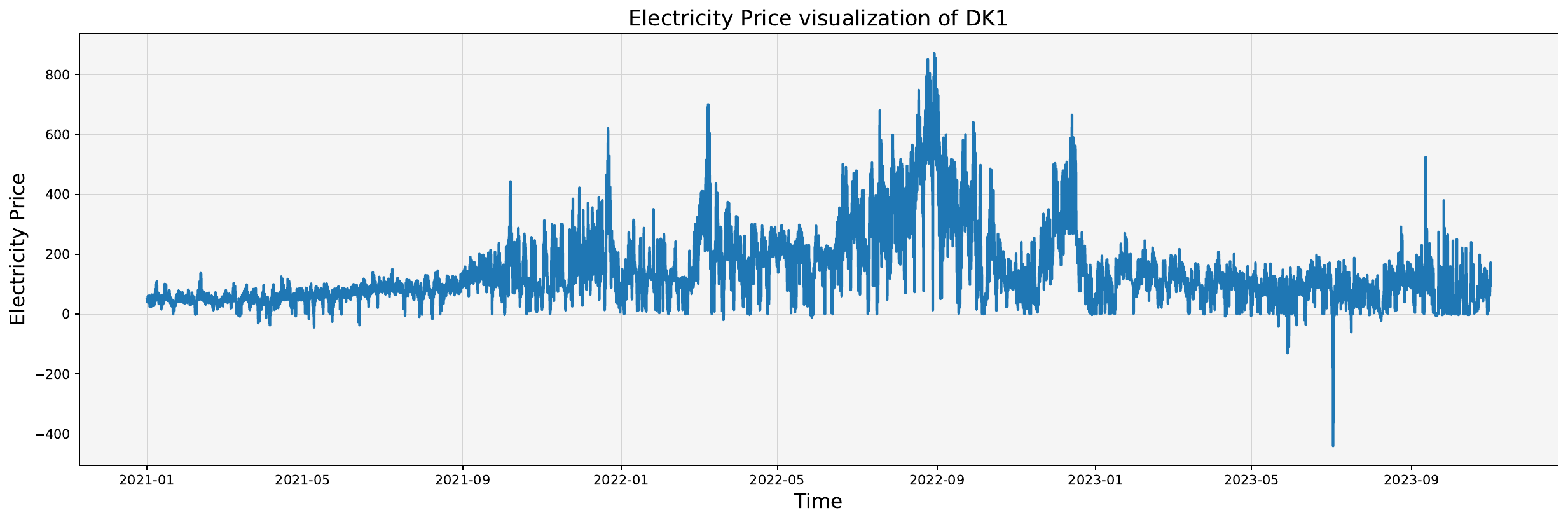}
\centering
\caption{Visualization of the electricity price series in the DK1 dataset.}
\label{DK1}
\end{figure}

\begin{figure}[htb]
\centering
\includegraphics[width=\textwidth]{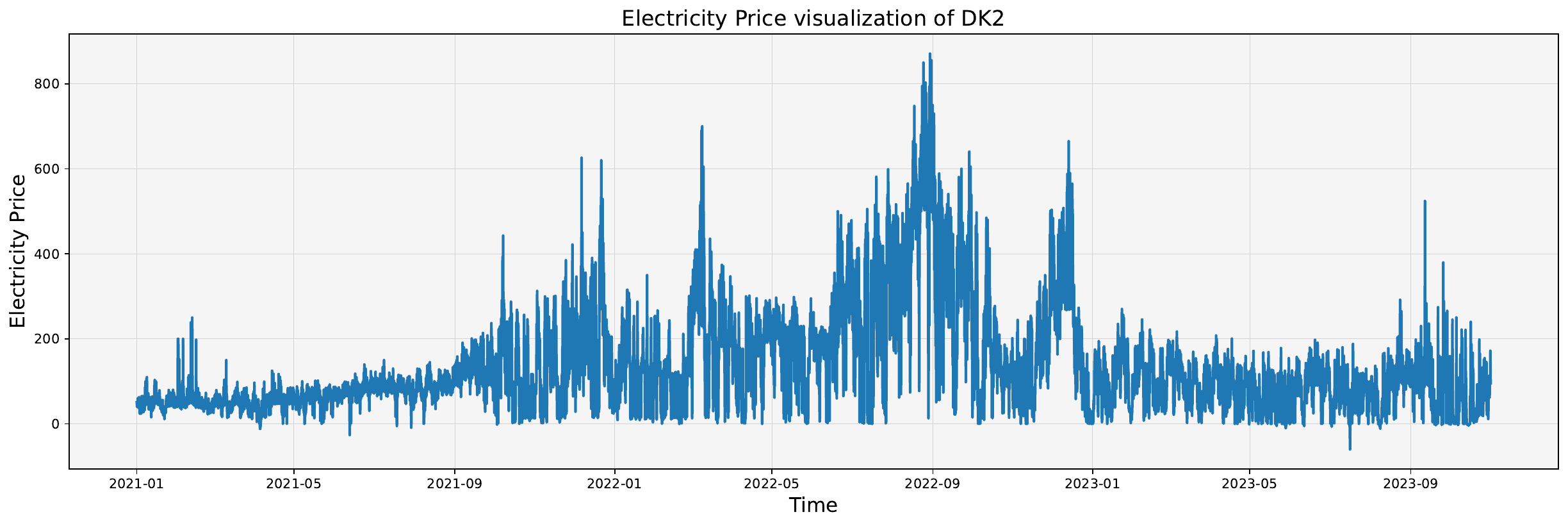}
\centering
\caption{Visualization of the electricity price series in the DK2 dataset.}
\label{DK2}
\end{figure}

Figure \ref{DK1} and Figure \ref{DK2} show the electricity price series, named DK1 and DK2, in the electricity markets of two regions in Denmark. We collect this data from Energinet\footnote{\href{https://en.energinet.dk/}{https://en.energinet.dk/}}, which is the Danish national transmission system operator for electricity and natural gas and it is an independent public enterprise owned by the Danish state under the Ministry of Climate and Energy. In addition to the electricity price data, we also provide corresponding data like load consumption and renewable energy output data including onshore wind power, offshore wind power, and PV, as our auxiliary variables.

\subsection{Renewable Energy Dataset}

\begin{figure}[htb]
\centering
\includegraphics[width=\textwidth]{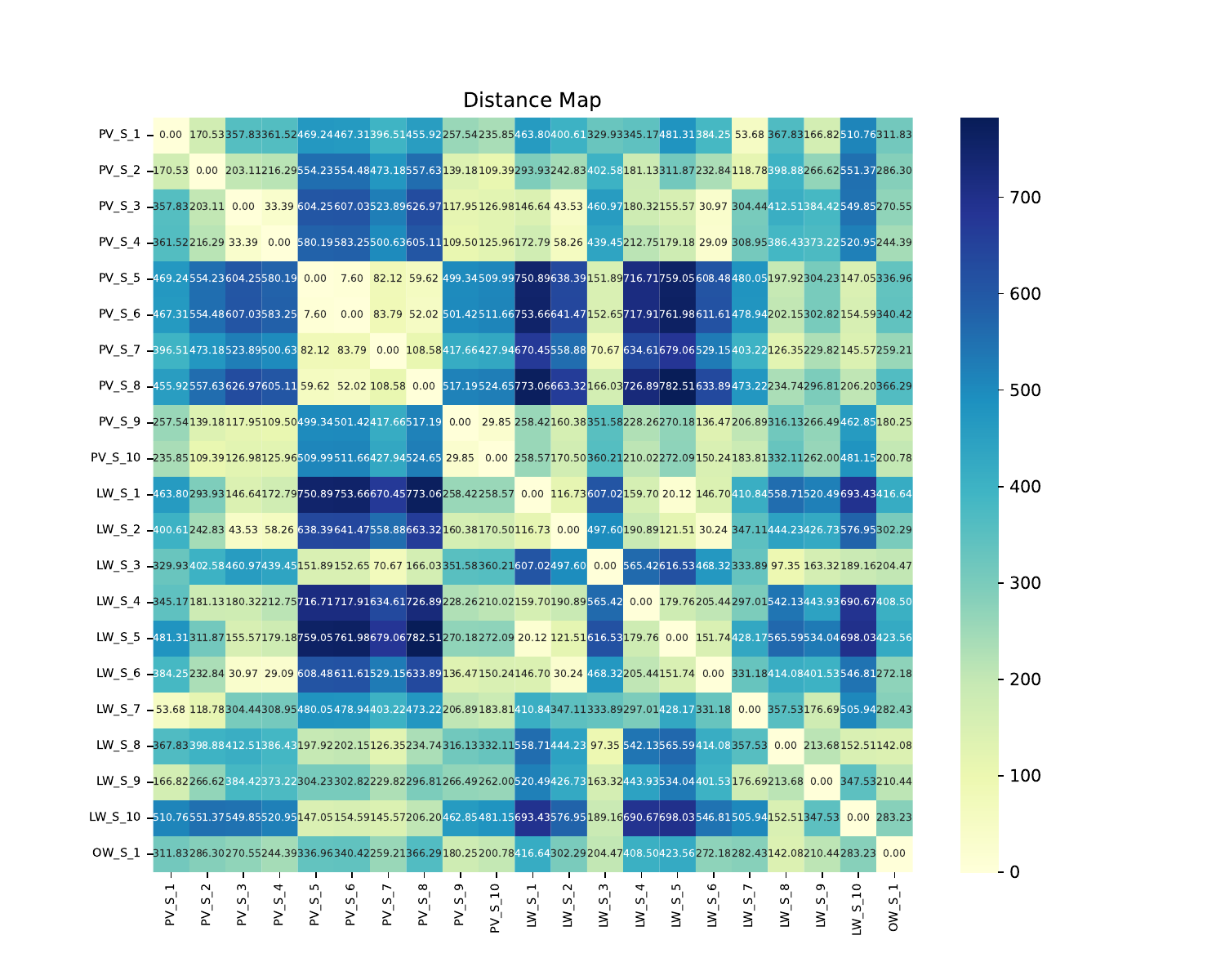}
\centering
\caption{Visualization of the distance between different station (KM).}
\label{distance}
\end{figure}

We have collected and released a new dataset on renewable energy generation, which includes onshore wind power (represented by LW), offshore wind power (represented by OW), and photovoltaics (represented by PV). At the same time, we also divide the renewable energy output data into station level (represented by S), city level (represented by C), and regional level (represented by R) according to geographical scale from small to large. Note that the data at the station level includes \textbf{real-time measurement} of meteorological factors (see Table \ref{renewable_dataset_S}). For the sake of confidentiality, we do not directly disclose the specific longitude and latitude of each station but give their relative positional relationships in Figure \ref{distance}. Users can download this dataset in our provided code repository after publication.

\begin{table}[htb]\Huge
\centering
\caption{Datasets in the renewable energy forecasting archive.}
\renewcommand{\arraystretch}{1.5}
\setlength{\tabcolsep}{4.5pt}
\resizebox{\textwidth}{!}{
\begin{tabular}{ccc}
\hline
\textbf{Data}                & \textbf{Installed Capacity (MW)} & \textbf{External Variables}                         \\ \hline
PV\_S\_1                     & 20                              & Irradiance($W/m^2$), Temperature(°C)                    \\
PV\_S\_2                     & 46                              & Irradiance($W/m^2$), Temperature(°C)                    \\
PV\_S\_3                     & 20                              & Irradiance($W/m^2$), Temperature(°C)                    \\
PV\_S\_4                     & 114.47                          & Irradiance($W/m^2$), Temperature(°C)                    \\
PV\_S\_5                     & 100                             & Irradiance($W/m^2$), Temperature(°C)                    \\
PV\_S\_6                     & 50                              & Irradiance($W/m^2$), Temperature(°C)                    \\
PV\_S\_7                     & 30                              & Irradiance($W/m^2$), Temperature(°C)                    \\
PV\_S\_8                     & 120                             & Irradiance($W/m^2$), Temperature(°C)                    \\
PV\_S\_9                     & 99                              & Irradiance($W/m^2$), Temperature(°C)                    \\
PV\_S\_10                    & 100                             & Irradiance($W/m^2$), Temperature(°C)                    \\ \hline
LW\_S\_1                     & 48                              & Wind speed($m/s$), Wind direction                     \\
LW\_S\_2                     & 40                              & Wind speed($m/s$), Wind direction                     \\
LW\_S\_3                     & 49.5                            & Wind speed($m/s$), Wind direction                     \\
LW\_S\_4                     & 36                              & Wind speed($m/s$), Wind direction                     \\
LW\_S\_5                     & 45.05                           & Wind speed($m/s$), Wind direction                     \\
LW\_S\_6                     & 66                              & Wind speed($m/s$), Wind direction                     \\
LW\_S\_7                     & 49.5                            & Wind speed($m/s$), Wind direction                     \\
LW\_S\_8                     & 49.5                            & Wind speed($m/s$), Wind direction                     \\
LW\_S\_9                     & 88                              & Wind speed($m/s$), Wind direction                     \\
LW\_S\_10                    & 39.5                            & Wind speed($m/s$)                                 \\ \hline
\multicolumn{1}{l}{OW\_S\_1} & 300                             & \multicolumn{1}{l}{Wind speed($m/s$), Wind direction} \\ \hline
\end{tabular}
}\label{renewable_dataset_S}
\end{table}

\section{Feature Analysis for Load Forecasting}
As we mentioned before, auxiliary variables have a significant impact on energy forecasting. Among them, temperature variables and calendar variables have a great impact on load forecasting. This is also recognized by the famous global energy forecasting competition. Based on this, the organizer developed a linear model called the vanilla model, which considers load, calendar variables, and temperature as the main variables and it serves as the benchmark for the forecasting competition \cite{hong2014global} and this is the intuition that we propose HT embedding. Therefore, in this section, we will visualize the relevant features in different levels of datasets to explore the relationship between load, calendar variables, and temperature.
\subsection{Temperature-load analysis}
\begin{figure}[htb]
\centering
\includegraphics[width=\textwidth]{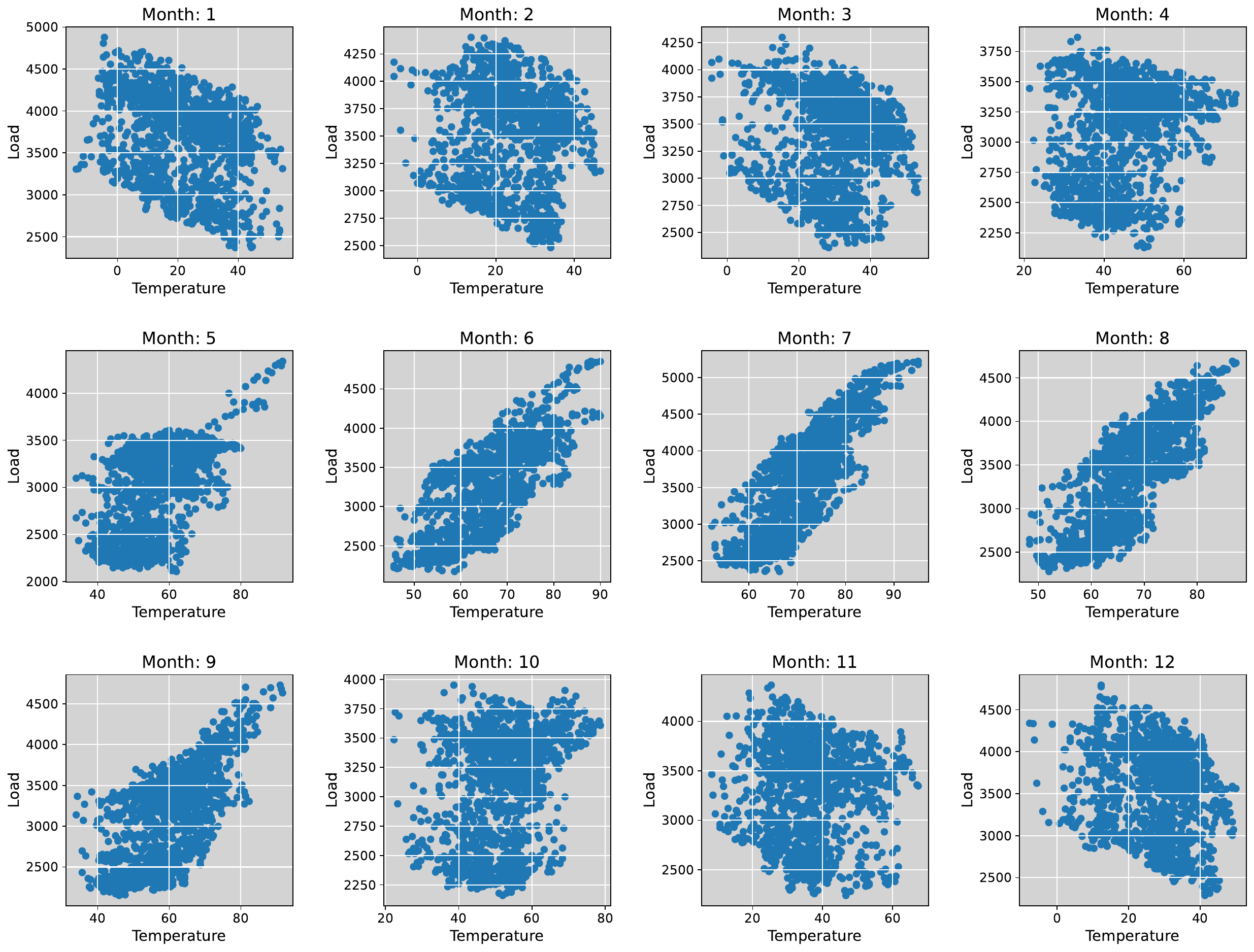}
\centering
\caption{Temperature-aggregated load scatter plots for 12 months.}
\label{agg_month}
\end{figure}

\begin{figure}[htb]
\centering
\includegraphics[width=\textwidth]{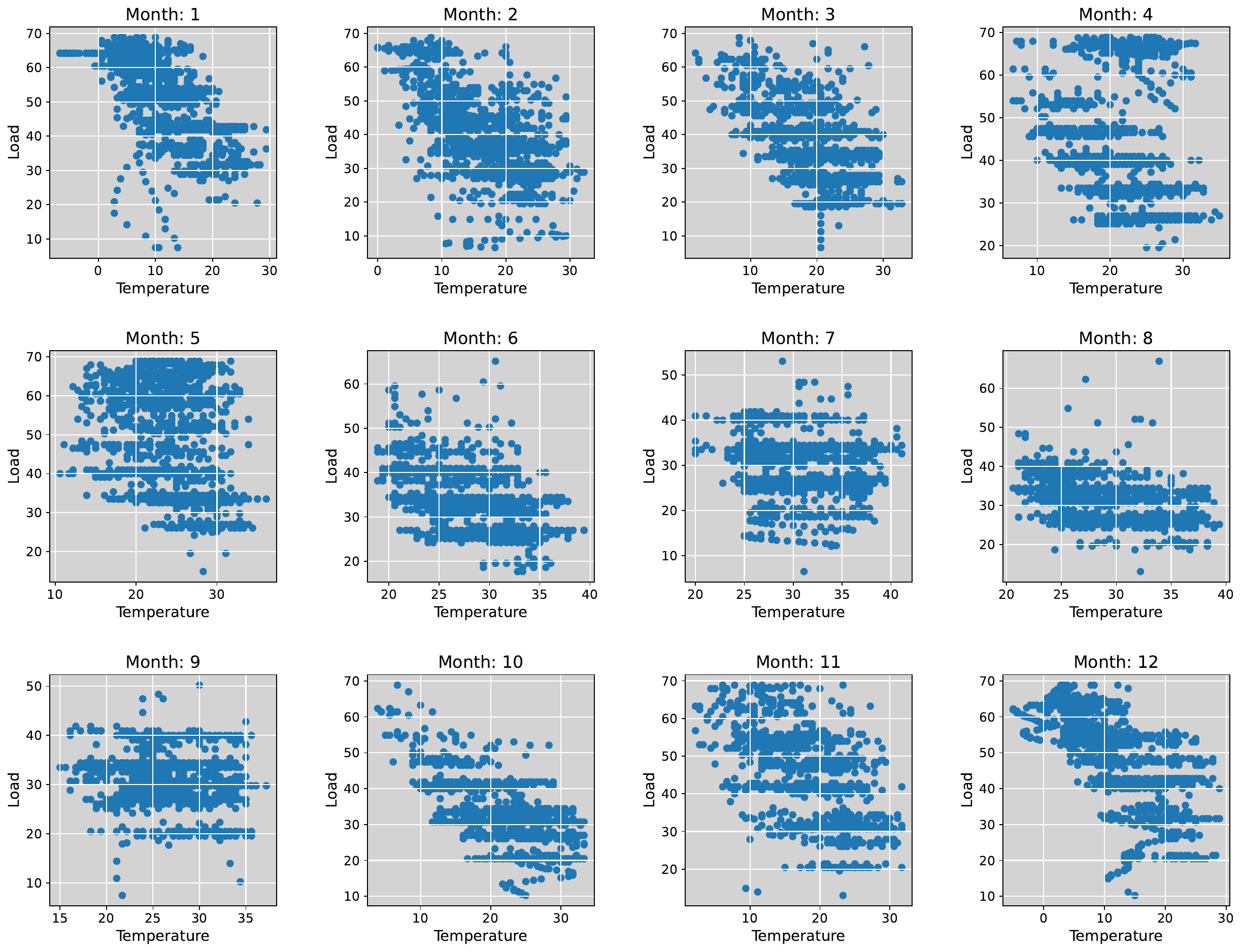}
\centering
\caption{Temperature-building load scatter plots for 12 months.}
\label{building_month}
\end{figure}

Figure \ref{agg_month} and \ref{building_month} show scatter plots of the relationship between temperature load at two different levels, respectively. The data at the aggregated level comes from the GEF14 competition, while the building level is randomly selected from the BDG2(Bull) dataset. It is worth noting that in the BDG2 dataset, each building has its corresponding attribute usage, such as educational facilities, office space, and so on. We divided the scatter plot of load and temperature into 12 blocks by month, with the aim of expanding the relationship between temperature and load to the ternary relationship between temperature, load, and calendar variables. Among them, we can consider calendar variables as indicators of seasons, months, workdays (weekends), and hours, and explore the temperature load relationships of different seasons (months, etc.). Similarly, we will mainly analyze the months here. 

From figure \ref{agg_month}, we can see that, in line with common sense, the relationship between load and temperature shows significant differences when in different months. From May to September, there is a significant positive correlation between load and temperature. Starting from October, the relationship between load and temperature gradually shifted from a significant positive correlation to an insignificant correlation. May to September is also a period of frequent high-temperature weather, indicating that when the temperature is high, there is a significant positive correlation between temperature and load. It is worth noting that this positive correlation is not always true. If we directly hand over the temperature variables to the model for modeling without processing, such changes in the relationship may cause confusion and ultimately lead to a decrease in forecasting performance.

When it comes to building-level load, figure \ref{building_month} shows that the uncertainty of the detected load is significantly greater than that of the aggregated load since each month presents different temperature-load relationships. As this is the load data from educational facilities, there may be classifications such as teaching days or rest days, as shown in the figure, where there is a clear phenomenon of fragmentation within multiple months.

In summary, the two levels of load exhibit different load-temperature relationships in different months. This situation also occurs in other time scales such as the hour. Therefore, to make the forecasting model understand this relationship correctly, it is necessary to consider calendar variables and temperature together.

\section{Package Usage}
\subsection{Asymmetric loss function}\label{simulated}
The ultimate goal of energy is to minimize subsequent scheduling costs, which are closely related to prediction errors. Inspired by \cite{zhang2022cost}, our package provides a piecewise linearized function to fit forecasting errors with other variables (which can be corresponding scheduling costs, etc.). At the same time, we also give an asymmetric loss function to replace the symmetric MSE loss function. Specifically, in actual power grid dispatch, the economic losses caused by forecasting values being less than the true values are often greater than the losses caused by forecasting values being greater than the true values. Therefore, we use a simple quadratic function to construct a piecewise generating function. 

Here, $\epsilon$ represents the Forecasting Error Percentage(FEP) $\epsilon=\frac{f\left(x\right)-y}{y}$. We first use this function to sample and obtain many data points and then use a smoothing spline, denoted as $s$, to fit them. To avoid many breakpoints, which may make it difficult to integrate into our forecasting framework as a loss function \cite{perperoglou2019review}, we use piecewise linearization to approximate the spline function. The selection of breakpoints can be based on the following formula\cite{berjon2015optimal},
$$\left\|s-L(\epsilon)\right\|_2 \leq \frac{\left(\int_{\epsilon_{\min }}^{\epsilon_{\max }} s^{\prime \prime}(\epsilon)^{\frac{2}{5}} d \epsilon\right)^{\frac{5}{2}}}{\sqrt{120} K^2},$$
where K is the number of breakpoints, and the integration interval we choose here is $(-0.15,0.15)$. By controlling the error between piecewise linear functions and spline functions, we can obtain an appropriate number of breakpoints. Here we control the error lower than 0.005. As for the location of the breakpoints, we first calculate the cumulative breakpoint distribution function according to \cite{de1978practical}, 
$$F\left(\epsilon_k\right)=\frac{\int_{\epsilon_{\min }}^{\epsilon_k}\left|s^{\prime \prime}(x)\right|^{2 / 5} d x}{\int_{\epsilon_{\min }}^{\epsilon_{\min }}\left|s^{\prime \prime}(x)\right|^{2 / 5} d x}.$$
The breakpoints $\left\{\epsilon_k\right\}_{k=1}^{K-1}$ will be placed such that each subinterval can contribute equally to the value of the cumulative breakpoint distribution function. To eventually integrate it into our forecasting framework as a loss function, we need to ensure that it is differentiable. Here we offer two options: either we can directly use the piecewise linearization function, as it is difficult for the value to accurately equal to a breakpoint in practical applications, or we can insert a quadratic function within the $0.000001$ distance before and after each breakpoint and obtain the parameters of the quadratic function by ensuring the continuity of the function and its first derivative at the two connections before and after. 

\begin{figure}[htb]
\centering
\includegraphics[width=\textwidth]{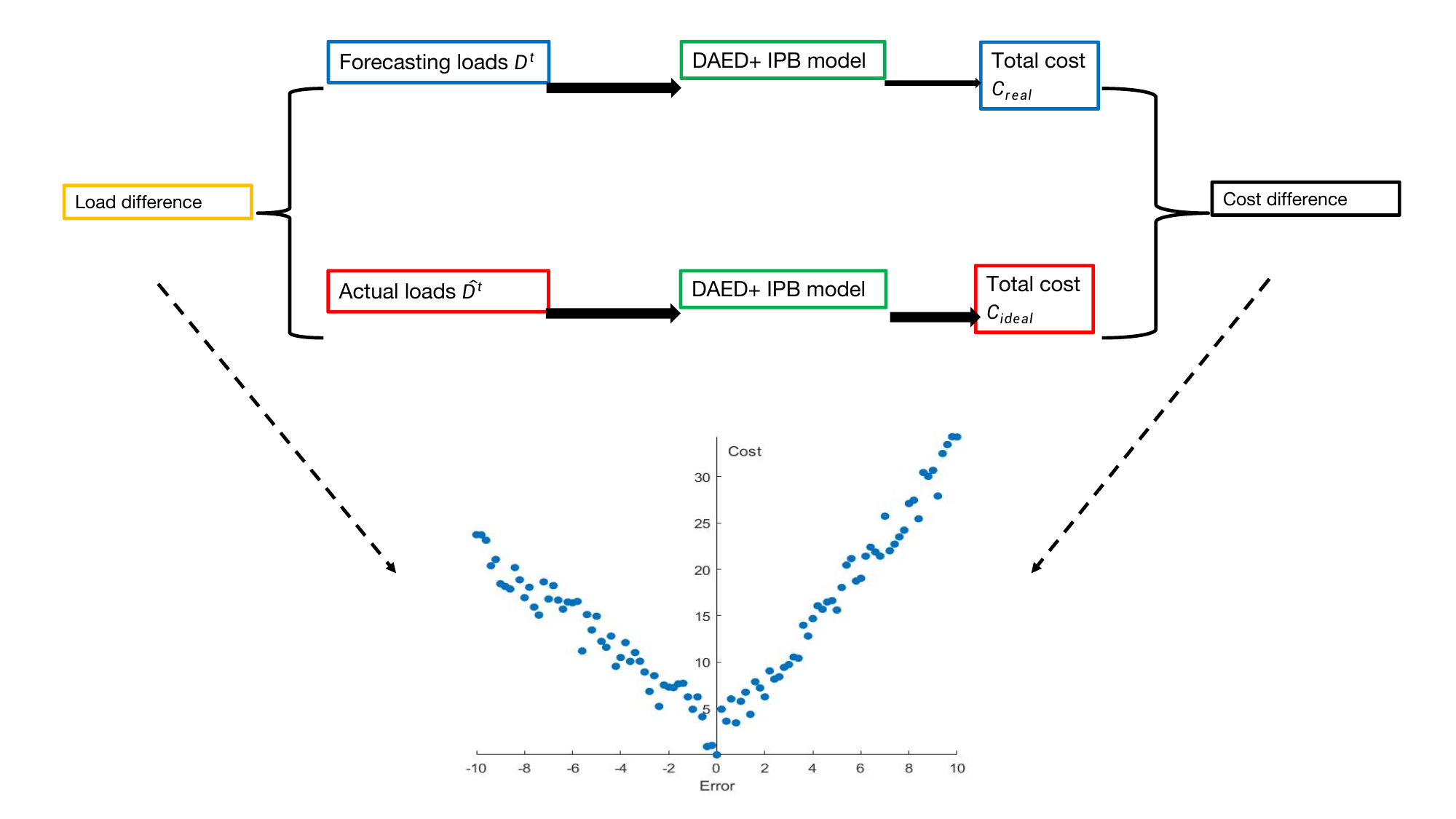}
\centering
\caption{Loss function generation (adapted from \cite{zhang2022cost}).}
\label{gener}
\end{figure}

\begin{figure}[htb]
\centering
\includegraphics[width=\textwidth]{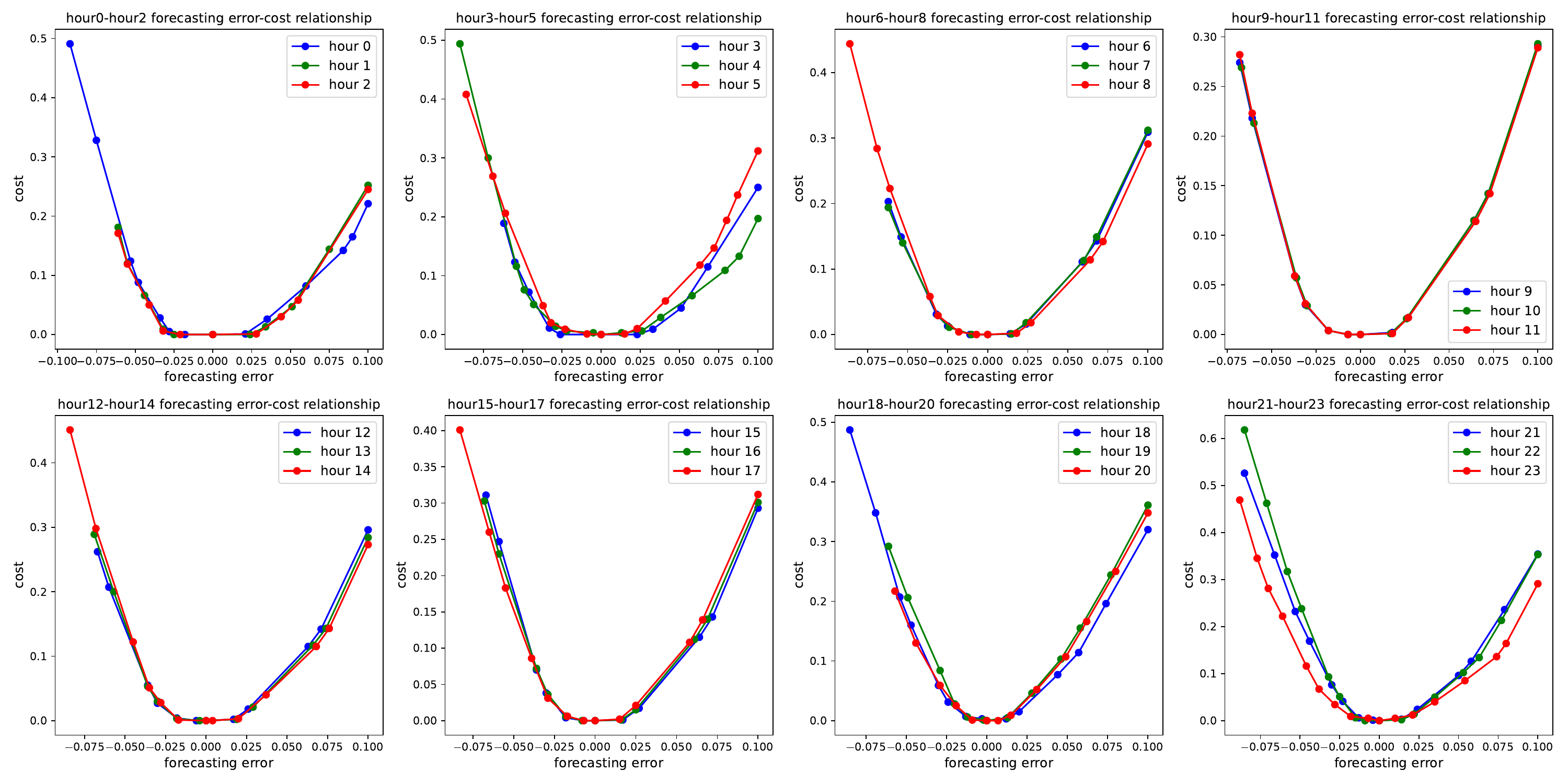}
\centering
\caption{Loss function visualization.}
\label{loss_function_all}
\end{figure}

As mentioned in section \ref{cus_loss}, we simulate an IEEE 30 bus system to get the relationship between forecasting error and dispatching cost. Fig. \ref{gener} shows the specific process of simulation. Here we mainly focus on two optimization problems DEAD and IPB. The mathematical definitions of these two can be found in \cite{zhang2022cost}. To solve these two optimization problems, we have provided the corresponding MATLAB code. Based on the above process, we have provided the relationship between the load forecasting error and the actual dispatching cost for each hour within 24 hours of a day, as shown in Fig \ref{loss_function_all}. All corresponding data is saved in the file "breakpoint$\_$new.mat" we provide and we can construct the corresponding loss function through a single line of code. Note that we use the simple linear version (may not be locally differentiable) and our experiment in the main text is conducted based on hour 8.

\begin{tcolorbox}[colback=lightgray!20, colframe=black, rounded corners=southeast, sharp corners=northwest]
\begin{lstlisting}[style=python]
loss_function = ContinuousPiecewiseLinearFunction(breakpoint)
\end{lstlisting}
\end{tcolorbox}

\begin{tcolorbox}[colback=lightgray!20, colframe=black, rounded corners=southeast, sharp corners=northwest]
\begin{lstlisting}[style=python]
loss_function = ContinuousPiecewiseFunction(break_point, overlap, linear_models, quadratic_models)
\end{lstlisting}
\end{tcolorbox}

\subsection{How to use the package}\label{package_usage}
\subsubsection{how to construct forecasting scenarios}

Our framework mainly constructs forecasting scenarios through the function "calculate$\_$scenario". Below, we will introduce how to prepare the input for this function separately.
\begin{tcolorbox}[colback=lightgray!20, colframe=black, rounded corners=southeast, sharp corners=northwest]
\begin{lstlisting}[style=python]
def calculate_scenario(data,
                       target,
                       methods_to_train,
                       horizon,
                       train_ratio,
                       feature_transformation,
                       time_stationarization,
                       datetime_features,
                       target_lag_selection,
                       ex_feature_selection,
                       post_processing_quantile,
                       post_processing_value,
                       evaluation_metrics,
                       ):
\end{lstlisting}
\end{tcolorbox}
\paragraph{data} The time series dataset and we need to input the data in Pandas DataFrame format.
\paragraph{target} The column name of the variable we need to forecast in the input DataFrame, and other columns should be treated as auxiliary variables.
\paragraph{methods$\_$to$\_$train} A list containing the forecasting methods we need to include.
\paragraph{horizon} The default is to make predictions 24 steps in advance and different time horizon forecasting can be made by adjusting this.
\paragraph{train$\_$ratio} The ratio of the division of the training and test sets.
\paragraph{feature$\_$transformation} The strategies used to stabilize the time series, including logarithmic transformation, data differentiation, and so on.
\paragraph{time$\_$stationarization} The division of the training and test sets.
\paragraph{datetime$\_$features} Define what calendar variables to consider, such as day, month, year, whether it is a holiday, etc.
\paragraph{target$\_$lag$\_$selection} Define how to select historical data for forecasting. In our default settings and our benchmark, we will select the past 24 time steps. In addition, we also provide a strategy for selecting highly correlated historical data based on the autocorrelation of the data.
\paragraph{ex$\_$feature$\_$selection} Define which auxiliary variables to use.
\paragraph{post$\_$processing$\_$quantile} Quantile-based forecasting may sometimes result in lower quantiles being greater than higher quantiles, and the main focus here is to rearrange them.
\paragraph{post$\_$processing$\_$value} To limit the final output result (like the restriction of installed capacity), such as forcing the forecasting result not to exceed a certain value.
\paragraph{evaluation$\_$metrics} We include various evaluation metrics for users to choose from, which can be referenced specifically from \ref{evaluation_metrics}.

\subsubsection{how to add new models}
Our package mainly focuses on quantile-based probabilistic forecasting while users can also implement deterministic models by changing 'MQ' to 'P' and choosing different loss functions in the following codes. To add new models, we need to make definitions for the new models.

\begin{tcolorbox}[colback=lightgray!20, colframe=black, rounded corners=southeast, sharp corners=northwest]
\begin{lstlisting}[style=python]
class MQPredictor(MultiQuantileRegressor):
    def __init__(self, quantiles: List[float],device,ex_model = None):
        super().__init__(
            X_scaler=StandardScaler(),
            y_scaler=StandardScaler(),
            quantiles=quantiles,
            ex_model = ex_model,
            device = device)
        

    def set_params(self, configs):
        self.model = models.pytorch.PytorchRegressor(
            model=my_MQuantile_model(configs).to(self.device),
            ex_model = define_ex_model(self.ex_model,configs),
            loss_function=pytorchtools.PinballLoss(self.quantiles,self.device),device =self.device)
        return self
\end{lstlisting}
\end{tcolorbox}

Note that ex$\_$model is a string that should be in [ex$\_$Linear, ex$\_$MLP, ex$\_$LSTM, ex$\_$Transformer, ex$\_$Mixer, ex$\_$HT] to represent different modules to handle the auxiliary variables (None means we don't use those modules.). In addition, users also need to provide specific details of the forecasting model, that is, how to handle the input of the model and ultimately convert it into output.
\begin{tcolorbox}[colback=lightgray!20, colframe=black, rounded corners=southeast, sharp corners=northwest]
\begin{lstlisting}[style=python]
class my_MQuantile_Model(nn.Module):
    def __init__(self,configs):
        super(MYQuantile_Model, self).__init__()
        #Code here
    def forward(self, *inputs):
        #Code here
        return output
\end{lstlisting}
\end{tcolorbox}

\section{Benchmark Evaluation}

\subsection{Hyper parameters setting}
In this section, we will introduce the hyperparameter settings in our energy forecasting archive.
Table \ref{methods} shows the parameter settings for non-deep learning methods. Here, KNNR represents quantile regression methods based on the K-nearest neighbor algorithm \cite{hastie2009elements}, RF represents quantile regression methods based on random forest \cite{meinshausen2006quantile}, and ERT represents quantile regression methods based on extreme random tree \cite{geurts2006extremely}. Note that since we have a lot of input features, we will perform PCA \cite{mackiewicz1993principal} dimensionality reduction before inputting the data into the model, and the dimension is selected as min$\{100,0.1\times input\_size\}$.

\begin{table}[htb]
\centering
\caption{Parameter settings for non-deep learning methods.}
\label{methods}
\renewcommand{\arraystretch}{1.2}
\setlength{\tabcolsep}{5pt}
\resizebox{\textwidth}{!}{
\begin{tabular}{cccccc} 
\hline
\multirow{2}{*}{\textbf{Method}} & \multicolumn{3}{c}{\textbf{Parameters}}                                          \\ \cline{2-4} 
                        & \textbf{N\_neighbors} & \textbf{N\_estimators} & \textbf{Quantiles}      \\ \hline
KNNR                                        & 5           & -             & 0.1$\sim$0.9 \\
RFR                                        & -            & 10           & 0.1$\sim$0.9 \\
ERT                                         & -            & 10           & 0.1$\sim$0.9 \\
\hline
\end{tabular}
}
\end{table}

Apart from those, we introduce several deep learning methods described in section \ref{sec3.2}. Tables \ref{Training} and \ref{tab:parameters} respectively show the hyperparameter settings of the training process and the network structure and parameters of the relevant deep learning methods. We divide the entire dataset into training and test sets at a ratio of 0.2, and then divide the training set into the final training and validation sets at a ratio of 0.2. To reduce the impact of neural network overfitting, we enable the early stop mechanism. Specifically, when the loss on the validation does not decrease for 5 epochs, we will stop training. As for the hyperparameter setting of the time series forecasting model, we mainly refer to \cite{wang2024tssurvey}. For more detailed settings, please refer to our code repository. In addition, note that as we mentioned in the main text, the ex-module we designed will implement a two-stage training. After the original time series forecasting model is trained, we will fix the time series model, concatenate its output and auxiliary variables, and perform additional fine-tuning steps. The fine-tuning epoch is set to 20, the patience remains unchanged at 5, and the learning rate will be selected from [1e-5, 1e-4] according to the validation.

\begin{table}[htb]\Huge
\centering
\caption{Training process parameters.}
\renewcommand{\arraystretch}{1}
\setlength{\tabcolsep}{4pt}
\resizebox{\textwidth}{!}{
\begin{tabular}{cccccccc}
\toprule
\multicolumn{6}{c}{\textbf{Parameters}} \\ \cmidrule(lr){1-7}
\textbf{Loss function} & \textbf{Validation ratio} & \textbf{Epochs} & \textbf{Patience} & \textbf{Optimizer} & \textbf{Learning rate} & \textbf{Batch size} \\ \midrule
\begin{tabular}[c]{@{}c@{}}MSE\\  PinballLoss (0.1$\sim$0.9)\end{tabular} & 0.2 & 50 & 5 & torch.Adam & 5e-4 & 32 \\ \bottomrule
\end{tabular}
}\label{Training}
\end{table}

\begin{table}[htb]
\centering
\resizebox{\textwidth}{!}{
\begin{tabular}{c|c}
\hline
\textbf{Parameter} & \textbf{Value} \\
\hline
d\_model  & 256 \\
n\_layers  & 2 \\
moving\_avg  & 25 \\
embed (Transformer-based methods) & 'timeF' \\
dropout (Transformer-based methods) & 0.1 \\
e\_layers (Transformer-based methods) & 2 \\
d\_layers (Transformer-based methods) & 1 \\
factor (Transformer-based methods) & 3 \\
n\_heads (Transformer-based methods) & 8 \\
d\_ff (Transformer-based methods) & 2048 \\
activation (Transformer-based methods) & 'gelu' \\
hidRNN (LSTNet) & 64 \\
hidCNN (LSTNet) & 32 \\
hidSkip (LSTNet) & 32 \\
CNN\_kernel (LSTNet) & 3 \\
skip (LSTNet) & 1 \\
highway\_window (LSTNet) & 0 \\
distil (Informer) & True \\
p\_hidden\_dims (NSTransformer) & [256, 256] \\
p\_hidden\_layers (NSTransformer) & 2 \\
seg\_len (SegRNN) & 24 \\
down\_sampling\_window (TimeMixer) & 2 \\
down\_sampling\_layers (TimeMixer) & 3 \\
channel\_independence (TimeMixer) & 1 \\
decomp\_method (TimeMixer) & 'moving\_avg' \\
use\_norm (TimeMixer) & 1 \\
down\_sampling\_method (TimeMixer) & 'avg' \\
top\_k (TimesNet) & 5 \\
num\_kernels (TimesNet) & 6 \\
Times\_Net\_d\_ff (TimesNet) & 512 \\
normalize\_before (TsMixer) & True \\
norm\_type (TsMixer) & 'batch' \\
num\_blocks (TsMixer) & 2 \\
ff\_dim (TsMixer) & d\_model \\
activation\_fn (TsMixer) & 'gelu' \\
static\_channels (TSMixeeExt) & 1 \\
hidden\_channels (TSMixeeExt) & 64 \\
kernel\_size (Wavenet) & 9 \\
Nl (Wavenet) & 5 \\
patch\_len (TimeXer) & 16 \\
features (TimeXer) & 'M' \\
use\_amp (WPMixer) & False \\
ex\_neurons (ex modules) & d\_model \\
\hline
\end{tabular}
}
\caption{Parameters setting of the time series forecasting models}
\label{tab:parameters}
\end{table}

\subsection{Evaluation metrics}\label{evaluation_metrics}
To evaluate the forecasting performance of different methods in our forecasting archive, we will introduce many evaluation metrics, which are divided into metrics for deterministic forecasting and metrics for probabilistic forecasting. It is worth noting that not all metrics are used to directly distinguish forecasting performance, and some of them may be used to describe the shape of probabilistic forecasting, thereby more comprehensively presenting the forecasting characteristics of different models. We will provide a detailed introduction below.

\subsubsection{Deterministic forecasting evaluation}
Similar to \cite{godahewa2021monash}, we adopt \textbf{4 metrics} that are widely used to evaluate the results of deterministic forecasting, and they are MAPE (Mean Absolute Percentage Error), MASE (Mean Absolute Scaled Error), RMSE (Root Mean Squared Error), and MAE (Root Mean Squared Error) respectively. Their mathematical definitions are listed below, note that $\{y_t\}_{t=1}^n$ represents the actual value and $\{F_t\}_{t=1}^n$ represents the predicted one.

\begin{itemize}
  \item \textbf{MAPE}. MAPE is a metric of forecasting accuracy that calculates the average percentage of forecasting error for all data points. The smaller the value of MAPE, the higher the forecasting accuracy. Due to its percentage error, it can be used to compare forecasting performance at different scales. However, MAPE may result in an infinite or very large error percentage for zero or near zero actual values. The formal definition of MAPE is given below
  \begin{equation*}
\text{MAPE} = \frac{1}{n} \sum_{t=1}^{n} \left| \frac{y_t - F_t}{y_t} \right| \times 100\%.
\end{equation*}
  \item \textbf{MASE}. MASE is a scale-independent error measure that calculates errors by comparing the forecasting error with the average absolute first-order difference of the actual value sequence. The advantage of MASE is that it is not affected by the size of actual values, so it is more robust for forecasting problems of different sizes. The formal definition of MASE is given below
  \begin{equation*}
\text{MASE} = \frac{1}{n} \sum_{t=1}^{n} \frac{\left| y_t - F_t \right|}{\frac{1}{n-1} \sum_{t=1}^{n} \left| y_{t} - y_{t-1} \right|}.
\end{equation*}
  \item \textbf{RMSE}. RMSE is a commonly used measure of forecasting error that calculates the square root of the average of the sum of squares of forecasting errors for all data points. The smaller the value of RMSE, the higher the forecasting accuracy. It is sensitive to outliers, which may lead to large forecasting errors. However, RMSE has good interpretability because its units are the same as the actual and predicted values. The formal definition of RMSE is given below
  \begin{equation*}
\text{RMSE} = \sqrt{\frac{1}{n} \sum_{t=1}^{n} (y_t - F_t)^2}.
\end{equation*}
 \item \textbf{MAE}. MAE is also a commonly used measure of forecasting error, which calculates the average of the absolute value of forecasting errors for all data points. The smaller the value of MAE, the higher the forecasting accuracy. Compared with RMSE, MAE is less sensitive to outliers, so it may be more robust in the case of outliers. The formal definition of MAE is given below
 \begin{equation*}
\text{MAE} = \frac{1}{n} \sum_{t=1}^{n} \left| y_t - F_t \right|.
\end{equation*}
\end{itemize}

\subsubsection{Probabilistic forecasting evaluation}

Compared to point forecasting, probabilistic forecasting can provide more information. Therefore, we can evaluate the results of probabilistic forecasting from more aspects. We have summarized a total of \textbf{11 metrics} to comprehensively evaluate the load probabilistic forecasting results and list them below. Note that we will also perform matrix visualization based on some metrics to help users better evaluate different prediction models.

\begin{itemize}

\item \textbf{CoverageError (CE)}. CoverageError is a method of measuring the quality of forecasting intervals, which measures the difference between the proportion of actual observations falling within the forecasting interval and the expected coverage rate. A smaller CoverageError indicates that the forecasting interval captures actual observations more accurately. Here, $L_t$ and $U_t$ represent the lower and upper bound of the forecasting interval while $UB$ and $LB$ respectively represent the upper and lower bounds of the interval we want. It is worth noting that when we visualize it, we call it ReliabilityMatrix. Specifically, we first divide the quantiles into the upper half and the lower half with 0.5 as the boundary. And perform pairwise combinations to obtain different nominal coverage rates as the horizontal axis, while the vertical axis represents the actual coverage rate.
\begin{equation*}
CE = \frac{1}{n} \sum_{t=1}^{n} (I(L_t \leq y_t \leq U_t) - (UB - LB)).
\end{equation*}

\item \textbf{Winkler Score (WS)}. Winkler Score (WS) is a metric that measures the quality of forecasting intervals. The forecasting interval is the forecasting range for future observations, usually represented by a lower bound and an upper bound. Winkler Score is used to evaluate whether the forecasting interval accurately captures actual observations, taking into account the width of the interval. A lower Winkler Score indicates better forecasting interval quality. Here, the symbols used are the same as CE while $\delta=U_t-L_t$. Similar to CE, in the corresponding visualization matrix, the abscissa should be different nominal coverage rates, and for a central (1-$\alpha$ )$\%$ forecasting interval, it is defined as follows:
  \begin{equation*}
W S_{a, t}= \begin{cases}\delta, & L_t \leq y_t \leq U_t. \\ \delta+\frac{2\left(y_t-U_t\right)}{\alpha}, & y_t>U_t. \\ \delta+\frac{2\left(L_t-y_t\right)}{\alpha}, & y_t<L_t.\end{cases}
\end{equation*}

\item \textbf{Pinball Loss (PL)}. Pinball Loss considers the difference between the forecasting value and the actual observation value, and weights the error based on whether the forecasting value falls on the side of the actual observation value (above or below). This enables Pinball Loss to capture the uncertainty in probabilistic forecasting and assign different weights to symmetric errors in loss calculations. A lower Pinball Loss indicates a smaller error between probabilistic forecasting and actual observations. Here, $L_\tau$ represents the Pinball Loss at the quantile $\tau$ and $\hat{y}_{\tau,t}$ is the forecasting value of corresponding time and quantile. In our setting, we consider the sum of 9 quantiles from 0.1 to 0.9, and it is defined as follows:
\begin{equation*}
PL=\frac{1}{n_\tau \cdot n} \sum_{t=1}^n \sum_{i=1}^{n_\tau} L_\tau\left(\hat{y}_{\tau,t}, y_t\right).
\end{equation*}

\item \textbf{RampScore (RS)}. RampScore measures the consistency of the slope (i.e. increasing or decreasing trend) between the forecasting sequence and the actual observation sequence. Firstly, we use the Swing Door compression algorithm \cite{khan2020impacts} to compress the forecasting sequence and the observed sequence, and then calculate the first-order difference values of these two sequences separately. Finally, we calculate the absolute difference between the first-order difference values of the two sequences and take the average to obtain the RampScore. A lower RampScore indicates that the model is more capable of capturing trends in sequence changes. Here, we calculate RampScore for 9 quantiles from 0.1 to 0.9.

\item \textbf{CalibrationError}. CalibrationError \cite{chung2021beyond} mainly evaluates the accuracy of forecasting models in representing uncertainty. The CalibrationError represents the difference between the forecasting quantile and the actual quantile. A smaller CalibrationError means that the forecasting model has higher accuracy in representing uncertainty, while a larger calibration error means that the forecasting model has lower accuracy in representing uncertainty. In the visualization matrix, we show the proportion of the predicted value greater than the true value under different quantiles. The closer the forecasting method is to the line y=x, the better the performance will be.

\end{itemize}

In addition to the metrics mentioned above, we also provide many other metrics. Although we will not present each of them in detail here, interested users can easily visualize them with the open-source code we provide. These evaluation metrics include IntervalWidth, QuantileCrossing, BoundaryCrossing, Skewness, Kurtosis, and QuartileDispersion. Among them, IntervalWidth calculates the width of probabilistic forecasting intervals given by different methods while QuantileCrossing gives the ratio of any two quantiles in which the predicted value of the lower quantile is greater than the predicted value of the higher quantile. BoundaryCrossing calculates the probability that the true value falls outside the forecasting range. Skewness and Kurtosis are metrics that describe the shape of a probability distribution. As for QuartileDispersion, its detailed description can be found in \cite{bonett2006confidence}.

\subsection{Evaluation results}\label{evaluation_result}

\begin{figure}[htb]
\centering
\includegraphics[width=\textwidth]{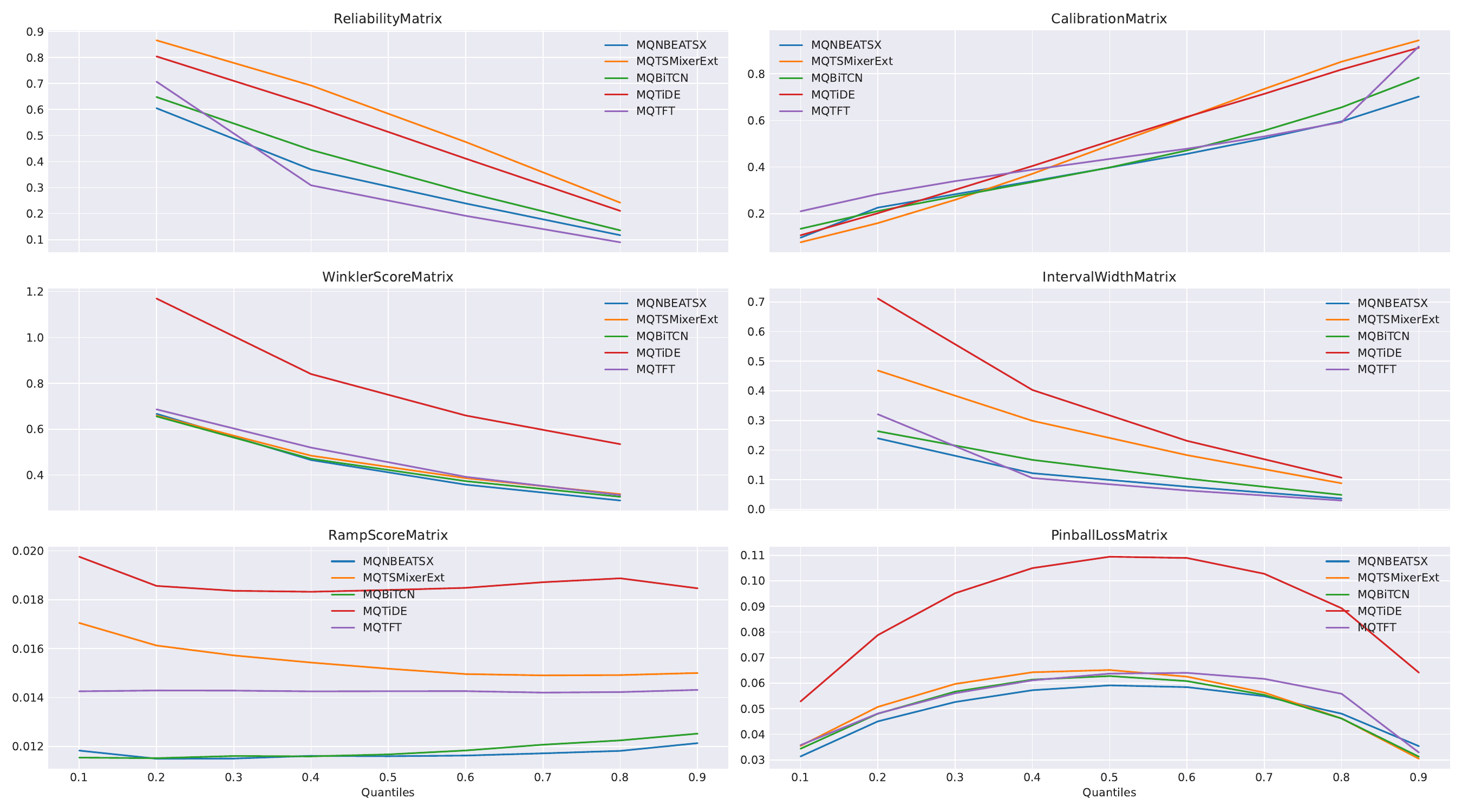}
\centering
\caption{Visualization of some of the probabilistic forecasting metrics.}
\label{fig:visualization}
\end{figure}

In this section, table\ref{fig:Forecasting_results} gives the overall performance while tables \ref{fig:aggload_results}, \ref{fig:buload_results}, \ref{fig:onwind_results}, \ref{fig:offwind_results}, \ref{fig:PV_results}, and \ref{fig:price_results} give the detailed results on the 21 datasets we collected in MAE and PinballLoss, as well as their total time consumption to make a direct comparison. Apart from that, figure \ref{fig:visualization} gives an example of how we visualize the metrics at different quantiles on GEF14 dataset. For other metrics, please refer to the implementation in the \href{https://github.com/Leo-VK/EnFoAV/}{repository} we provide.

\begin{table*}[htb]
\centering
\caption{Summary of the forecasting results. Every method runs 3 times and reports the average.}
\renewcommand{\arraystretch}{1.2}
\resizebox{0.95\textwidth}{!}{
\begin{tabular}{c|cccccccccccccccc}
\hline
\textbf{Method\textbackslash{}Dataset} & \multicolumn{2}{c}{\textbf{Aggregated Level Load}} & \multicolumn{2}{c}{\textbf{Building Level Load}} & \multicolumn{2}{c}{\textbf{Onshore Wind}} & \multicolumn{2}{c}{\textbf{Offshore Wind}} & \multicolumn{2}{c}{\textbf{Photovoltaic}} & \multicolumn{2}{c}{\textbf{Electricity Price}} & \multicolumn{2}{c}{\textbf{Average}} & \multicolumn{2}{c}{\textbf{Rank}} \\ \hline
-                                      & MAE$\downarrow$                  & Pinball Loss$\downarrow$                 & MAE$\downarrow$                  & Pinball Loss$\downarrow$                & MAE$\downarrow$               & Pinball Loss$\downarrow$            & MAE$\downarrow$               & Pinball Loss$\downarrow$             & MAE$\downarrow$               & Pinball Loss$\downarrow$            & MAE$\downarrow$                 & Pinball Loss$\downarrow$               & MAE$\downarrow$            & Pinball Loss$\downarrow$          & MAE Rank$\uparrow$          & Pinball Loss Rank$\uparrow$        \\ \hline
MLP                                    & 0.394                & 0.154                       & 0.420               & 0.164                      & 0.462            & 0.178                  & 0.455            & 0.176                   & 0.339            & 0.120                  & 0.240              & 0.091                     & 0.385         & 0.147                & 19          & 4                   \\
LSTM                                   & 0.546                & 0.212                       & 0.494               & 0.197                      & 0.543            & 0.233                  & 0.521            & 0.229                   & 0.328            & 0.127                  & 0.295              & 0.102                     & 0.454         & 0.183                & 25          & 26                  \\
\hline
\end{tabular}
}
\label{fig:Forecasting_results}
\end{table*}

\begin{table*}[htb]
\centering
\caption{Summary of the forecasting results of Aggregated Level Load. Every method runs 3 times and reports the average.}
\renewcommand{\arraystretch}{1.1}
\resizebox{0.95\textwidth}{!}{
\begin{tabular}{c|cccccccccccccc}
\hline
\textbf{Method\textbackslash{}Dataset} & \multicolumn{2}{c}{\textbf{Covid19}} & \multicolumn{2}{c}{\textbf{GEF12}} & \multicolumn{2}{c}{\textbf{GEF14}} & \multicolumn{2}{c}{\textbf{GEF17}} & \multicolumn{2}{c}{\textbf{PDB}} & \multicolumn{2}{c}{\textbf{Spain}} & \multicolumn{2}{c}{\textbf{Panama}} \\ \hline
-                                      & MAE               & Pinball Loss     & MAE              & Pinball Loss    & MAE              & Pinball Loss    & MAE              & Pinball Loss    & MAE             & Pinball Loss   & MAE              & Pinball Loss    & MAE              & Pinball Loss     \\ \hline
MLP                                    & 0.359$\pm$0.02    & 0.145$\pm$0.01   & 0.987$\pm$0.09   & 0.416$\pm$0.05  & 0.178$\pm$0.01   & 0.068$\pm$0.00   & 0.240$\pm$0.01   & 0.093$\pm$0.00   & 0.182$\pm$0.01  & 0.069$\pm$0.00  & 0.362$\pm$0.00    & 0.111$\pm$0.00   & 0.449$\pm$0.01   & 0.175$\pm$0.01   \\
LSTM                                   & 0.662$\pm$0.19    & 0.224$\pm$0.04   & 0.881$\pm$0.07   & 0.408$\pm$0.03  & 0.289$\pm$0.20    & 0.127$\pm$0.09  & 0.274$\pm$0.01   & 0.110$\pm$0.00    & 0.353$\pm$0.23  & 0.125$\pm$0.08 & 0.727$\pm$0.18   & 0.203$\pm$0.09  & 0.635$\pm$0.09   & 0.284$\pm$0.02   \\
CNN                                    & 0.328$\pm$0.02    & 0.093$\pm$0.00    & 1.277$\pm$0.27   & 0.811$\pm$0.32  & 0.216$\pm$0.01   & 0.090$\pm$0.01   & 0.268$\pm$0.01   & 0.114$\pm$0.03  & 0.224$\pm$0.01  & 0.089$\pm$0.00  & 0.460$\pm$0.02    & 0.163$\pm$0.00   & 0.538$\pm$0.15   & 0.161$\pm$0.06   \\
Transformer                            & 0.394$\pm$0.03    & 0.190$\pm$0.01    & 0.876$\pm$0.06   & 0.438$\pm$0.03  & 0.194$\pm$0.01   & 0.084$\pm$0.00   & 0.248$\pm$0.01   & 0.110$\pm$0.01   & 0.200$\pm$0.01    & 0.083$\pm$0.01 & 0.400$\pm$0.01     & 0.138$\pm$0.01  & 0.561$\pm$0.03   & 0.262$\pm$0.02   \\
\hline
\end{tabular}
}
\label{fig:aggload_results}
\end{table*}

\begin{table*}[htb]
\centering
\caption{Summary of the forecasting results of Building Level Load. Every method runs 3 times and reports the average.}
\renewcommand{\arraystretch}{1.1}
\resizebox{0.75\textwidth}{!}{
\begin{tabular}{c|cccccc}
\hline
\textbf{Method\textbackslash{}Dataset} & \multicolumn{2}{c}{\textbf{Hog}} & \multicolumn{2}{c}{\textbf{Bull}} & \multicolumn{2}{c}{\textbf{Cockatoo}} \\ \hline
-                                      & MAE             & Pinball Loss   & MAE             & Pinball Loss    & MAE               & Pinball Loss      \\ \hline
MLP                                    & 0.347$\pm$0.01  & 0.132$\pm$0.00  & 0.611$\pm$0.01  & 0.244$\pm$0.00   & 0.302$\pm$0.01    & 0.115$\pm$0.01    \\
LSTM                                   & 0.402$\pm$0.01  & 0.164$\pm$0.01 & 0.684$\pm$0.02  & 0.282$\pm$0.01  & 0.395$\pm$0.15    & 0.146$\pm$0.05    \\
CNN                                    & 0.347$\pm$0.01  & 0.139$\pm$0.01 & 0.585$\pm$0.02  & 0.258$\pm$0.02  & 0.316$\pm$0.02    & 0.140$\pm$0.05     \\
Transformer                            & 0.335$\pm$0.01  & 0.146$\pm$0.01 & 0.572$\pm$0.01  & 0.258$\pm$0.01  & 0.316$\pm$0.02    & 0.139$\pm$0.01    \\
LSTNet                                 & 0.466$\pm$0.01  & 0.185$\pm$0.00  & 0.720$\pm$0.05   & 0.275$\pm$0.01  & 0.305$\pm$0.01    & 0.119$\pm$0.01    \\
WaveNet                                & 0.413$\pm$0.02  & 0.205$\pm$0.01 & 0.656$\pm$0.02  & 0.309$\pm$0.02  & 0.352$\pm$0.03    & 0.182$\pm$0.01    \\
\hline
\end{tabular}
}
\label{fig:buload_results}
\end{table*}

\begin{table*}[htb]
\centering
\caption{Summary of the forecasting results of Onshore Wind. Every method runs 3 times and reports the average.}
\renewcommand{\arraystretch}{1.1}
\resizebox{0.75\textwidth}{!}{
\begin{tabular}{c|cccccc}
\hline
\textbf{Method\textbackslash{}Dataset} & \multicolumn{2}{c}{\textbf{Station}} & \multicolumn{2}{c}{\textbf{City}} & \multicolumn{2}{c}{\textbf{Region}} \\ \hline
-                                      & MAE               & Pinball Loss     & MAE             & Pinball Loss    & MAE              & Pinball Loss     \\ \hline
MLP                                    & 0.570$\pm$0.01     & 0.216$\pm$0.00    & 0.480$\pm$0.01   & 0.187$\pm$0.01  & 0.337$\pm$0.01   & 0.131$\pm$0.00    \\
LSTM                                   & 0.662$\pm$0.02    & 0.285$\pm$0.01   & 0.559$\pm$0.01  & 0.239$\pm$0.01  & 0.407$\pm$0.02   & 0.175$\pm$0.01   \\
CNN                                    & 0.548$\pm$0.02    & 0.239$\pm$0.02   & 0.465$\pm$0.02  & 0.210$\pm$0.03   & 0.333$\pm$0.01   & 0.131$\pm$0.00    \\
Transformer                            & 0.547$\pm$0.01    & 0.248$\pm$0.01   & 0.468$\pm$0.02  & 0.215$\pm$0.01  & 0.346$\pm$0.01   & 0.158$\pm$0.01   \\
LSTNet                                 & 0.815$\pm$0.01    & 0.313$\pm$0.01   & 0.669$\pm$0.01  & 0.262$\pm$0.00   & 0.468$\pm$0.01   & 0.194$\pm$0.04   \\
WaveNet                                & 0.768$\pm$0.02    & 0.352$\pm$0.02   & 0.671$\pm$0.02  & 0.322$\pm$0.01  & 0.608$\pm$0.02   & 0.292$\pm$0.02   \\
\hline
\end{tabular}
}
\label{fig:onwind_results}
\end{table*}

\begin{table*}[htb]
\centering
\caption{Summary of the forecasting results of Offshore Wind. Every method runs 3 times and reports the average.}
\renewcommand{\arraystretch}{1.1}
\resizebox{0.75\textwidth}{!}{
\begin{tabular}{c|cccccc}
\hline
\textbf{Method\textbackslash{}Dataset} & \multicolumn{2}{c}{\textbf{Station}} & \multicolumn{2}{c}{\textbf{City}} & \multicolumn{2}{c}{\textbf{Region}} \\ \hline
-                                      & MAE               & Pinball Loss     & MAE             & Pinball Loss    & MAE              & Pinball Loss     \\ \hline
MLP                                    & 0.389$\pm$0.02    & 0.149$\pm$0.01   & 0.564$\pm$0.02  & 0.215$\pm$0.01  & 0.413$\pm$0.01   & 0.163$\pm$0.00    \\
LSTM                                   & 0.400$\pm$0.02      & 0.190$\pm$0.08    & 0.67$\pm$0.02   & 0.293$\pm$0.02  & 0.494$\pm$0.02   & 0.204$\pm$0.01   \\
CNN                                    & 0.395$\pm$0.02    & 0.159$\pm$0.01   & 0.519$\pm$0.01  & 0.203$\pm$0.00   & 0.403$\pm$0.01   & 0.162$\pm$0.00    \\
Transformer                            & 0.390$\pm$0.02     & 0.172$\pm$0.01   & 0.566$\pm$0.01  & 0.243$\pm$0.01  & 0.437$\pm$0.01   & 0.198$\pm$0.01   \\
LSTNet                                 & 0.389$\pm$0.02    & 0.158$\pm$0.01   & 0.818$\pm$0.02  & 0.319$\pm$0.00   & 0.637$\pm$0.02   & 0.256$\pm$0.00    \\
WaveNet                                & 0.692$\pm$0.06    & 0.344$\pm$0.04   & 0.943$\pm$0.03  & 0.451$\pm$0.02  & 0.827$\pm$0.03   & 0.424$\pm$0.04   \\
\hline
\end{tabular}
}
\label{fig:offwind_results}
\end{table*}

\begin{table*}[htb]
\centering
\caption{Summary of the forecasting results of PV. Every method runs 3 times and reports the average.}
\renewcommand{\arraystretch}{1.1}
\resizebox{0.75\textwidth}{!}{
\begin{tabular}{c|cccccc}
\hline
\textbf{Method\textbackslash{}Dataset} & \multicolumn{2}{c}{\textbf{Station}} & \multicolumn{2}{c}{\textbf{City}} & \multicolumn{2}{c}{\textbf{Region}} \\ \hline
-                                      & MAE               & Pinball Loss     & MAE             & Pinball Loss    & MAE              & Pinball Loss     \\ \hline
MLP                                    & 0.375$\pm$0.04    & 0.13$\pm$0.010    & 0.370$\pm$0.01   & 0.137$\pm$0.00   & 0.272$\pm$0.03   & 0.092$\pm$0.01   \\
LSTM                                   & 0.369$\pm$0.05    & 0.137$\pm$0.01   & 0.339$\pm$0.01  & 0.138$\pm$0.00   & 0.276$\pm$0.02   & 0.106$\pm$0.02   \\
CNN                                    & 0.496$\pm$0.09    & 0.188$\pm$0.05   & 0.531$\pm$0.05  & 0.206$\pm$0.03  & 0.414$\pm$0.06   & 0.182$\pm$0.06   \\
Transformer                            & 0.357$\pm$0.04    & 0.151$\pm$0.02   & 0.346$\pm$0.01  & 0.146$\pm$0.01  & 0.242$\pm$0.02   & 0.100$\pm$0.02     \\
LSTNet                                 & 0.404$\pm$0.05    & 0.159$\pm$0.03   & 0.361$\pm$0.01  & 0.138$\pm$0.00   & 0.270$\pm$0.02    & 0.104$\pm$0.01   \\
WaveNet                                & 0.776$\pm$0.08    & 0.327$\pm$0.07   & 0.754$\pm$0.06  & 0.305$\pm$0.02  & 0.734$\pm$0.07   & 0.288$\pm$0.04   \\
\hline
\end{tabular}
}
\label{fig:PV_results}
\end{table*}

\begin{table*}[htb]
\centering
\caption{Summary of the forecasting results of Electricity Price. Every method runs 3 times and reports the average.}
\renewcommand{\arraystretch}{1.1}
\resizebox{0.75\textwidth}{!}{
\begin{tabular}{c|cccc}
\hline
\textbf{Method\textbackslash{}Dataset} & \multicolumn{2}{c}{\textbf{DK1}} & \multicolumn{2}{c}{\textbf{DK2}} \\ \hline
-                                      & MAE               & Pinball Loss     & MAE             & Pinball Loss    \\ \hline
MLP                                    & 0.242$\pm$0.01    & 0.092$\pm$0.00    & 0.238$\pm$0.01  & 0.090$\pm$0.00    \\
LSTM                                   & 0.254$\pm$0.02    & 0.100$\pm$0.01     & 0.336$\pm$0.13  & 0.103$\pm$0.02  \\
CNN                                    & 0.214$\pm$0.01    & 0.086$\pm$0.00    & 0.224$\pm$0.01  & 0.087$\pm$0.00   \\
Transformer                            & 0.238$\pm$0.01    & 0.098$\pm$0.00    & 0.240$\pm$0.01   & 0.098$\pm$0.00   \\
LSTNet                                 & 0.230$\pm$0.01     & 0.090$\pm$0.00     & 0.235$\pm$0.01  & 0.091$\pm$0.00   \\
WaveNet                                & 0.310$\pm$0.02     & 0.144$\pm$0.01   & 0.446$\pm$0.05  & 0.163$\pm$0.01  \\
\hline
\end{tabular}
}
\label{fig:price_results}
\end{table*}

\begin{table*}[htb]
\centering
\caption{Total time consumption of different methods.}
\renewcommand{\arraystretch}{1.1}
\resizebox{0.75\textwidth}{!}{
\begin{tabular}{c|cc}
\hline
\textbf{Method} & \multicolumn{2}{c}{\textbf{Time(hour)}} \\ \hline
-               & Deterministic forecasting            & Probabilistic forecasting        \\ \hline
MLP$\times$5(6)             & 3.886          & 3.582               \\
LSTM$\times$5(6)              & 41.695         & 43.614              \\
CNN$\times$5(6)               & 5.530          & 5.613               \\
Transformer$\times$5(6)       & 25.488         & 39.516              \\
LSTNet$\times$5(6)            & 9.407          & 9.907               \\
WaveNet$\times$5(6)           & 7.250          & 8.590               \\
\hline
\end{tabular}
}
\begin{tablenotes}
\footnotesize
\item[1] $\times5(6)$ : This means that for these methods, we combine them with every ex-module, so there will be extra training time.
\end{tablenotes}
\label{tab:Tconsumption}
\end{table*}

\clearpage

\section{Documentation}
\textbf{Long-term preserve plan}: Currently, our relevant datasets and forecasting results are saved in the folder in a cloud service. This is mainly because we are still updating it, and the main direction is to add more fine-grained data related to smart meters. After our dataset is fully developed, we will apply for the relevant DOI for it.
\textbf{Author statement}: We confirm that the relevant dataset sources comply with relevant regulations and we bear all responsibility in case of violation of rights, etc., and confirmation of the data license.

\clearpage

\bibliographystyle{elsarticle-num} 
\bibliography{cas-refs}
\end{document}